\pdfoutput=1

\documentclass[11pt]{article}

\usepackage[final]{acl}

\usepackage{times}
\usepackage{latexsym}

\usepackage[T1]{fontenc}

\usepackage[utf8]{inputenc}

\usepackage{microtype}

\usepackage{inconsolata}

\usepackage{graphicx}

\usepackage[utf8]{inputenc} 
\usepackage[T1]{fontenc}    
\usepackage{hyperref}       
\usepackage{url}            
\usepackage{booktabs}       
\usepackage{amsfonts}       
\usepackage{nicefrac}       
\usepackage{microtype}      
\usepackage{xcolor}         
\usepackage{subfig}
\usepackage{times}
\usepackage{latexsym}
\usepackage{xcolor}
\usepackage{tabularx}
\usepackage{amssymb}
\usepackage{graphicx}
\usepackage{multicol}
\usepackage{hhline}
\usepackage{multirow}
\usepackage{booktabs}
\usepackage{amssymb}
\usepackage{pifont}
\usepackage{tcolorbox}
\newcommand{\xmark}{\ding{55}}%

\usepackage{mathtools}
\DeclarePairedDelimiterX{\infdivx}[2]{(}{)}{%
  #1\;\delimsize\|\;#2%
}
\newcommand{\infdiv}{D_{\mathrm{KL}}\infdivx}

\usepackage{listings}
\usepackage{adjustbox}
\usepackage{amsmath}
\usepackage{amssymb}
\usepackage{cleveref}
\usepackage{enumitem}

\usepackage{algorithm}
\usepackage{algpseudocode, algorithmicx}

\newcommand{\ours}{Text2Chart31}
\definecolor{gray}{HTML}{f1f3f5}

\tcbset{sharp corners,before skip = 0.2cm,after skip = 0.5cm,left = 1mm,right = 1mm,top = 0.5mm,bottom = 0.5mm,colback = gray,boxrule = 0.1pt}
\newtcolorbox{customBox}{}

\title{Text2Chart31: Instruction Tuning for Chart Generation \\with Automatic Feedback}

\author{
 \textbf{Fatemeh Pesaran zadeh\textsuperscript{1}} \quad
 \textbf{Juyeon Kim\textsuperscript{2}$^*$} \quad 
 \textbf{Jin-Hwa Kim\textsuperscript{1,3}} \quad
 \textbf{Gunhee Kim\textsuperscript{1}$^\dagger$}\quad
\\
 \textsuperscript{1}Seoul National University,
 \textsuperscript{2}KAIST AI,
 \textsuperscript{3}NAVER AI Lab
\\
\texttt{\small fatemehpesaran@vision.snu.ac.kr, juyeonkim@kaist.ac.kr,} \\
\texttt{\small j1nhwa.kim@navercorp.com, gunhee@snu.ac.kr} 
}
\newcommand{\correspondingfootnote}{
    \let\oldthefootnote=\thefootnote
    \renewcommand{\thefootnote}{}
    \footnotetext{$*$ Work done as an intern at Seoul National University}
    \footnotetext{$\dagger$ Corresponding author.}
    \let\thefootnote=\oldthefootnote
}

\begin{document}
\maketitle
\begin{abstract}
Large language models (LLMs) have demonstrated strong capabilities across various language tasks, notably through instruction-tuning methods. However, LLMs face challenges in visualizing complex, real-world data through charts and plots.
Firstly, existing datasets rarely cover a full range of chart types, such as 3D, volumetric, and gridded charts.
Secondly, supervised fine-tuning methods do not fully leverage the intricate relationships within rich datasets, including text, code, and figures. To address these challenges, we propose a hierarchical pipeline and a new dataset for chart generation.
Our dataset, Text2Chart31, includes 31 unique plot types referring to the Matplotlib library, with 11.1K tuples of descriptions, code, data tables, and plots. 
Moreover, we introduce a reinforcement learning-based instruction tuning technique for chart generation tasks without requiring human feedback.
Our experiments show that this approach significantly enhances the model performance, enabling smaller models to outperform larger open-source models and be comparable to state-of-the-art proprietary models in data visualization tasks.
We make the code and dataset available at \url{https://github.com/fatemehpesaran310/Text2Chart31}.
\end{abstract}

\correspondingfootnote

\section{Introduction}

\begin{figure}[t]
    \centering
    \includegraphics[width=1.0\columnwidth]{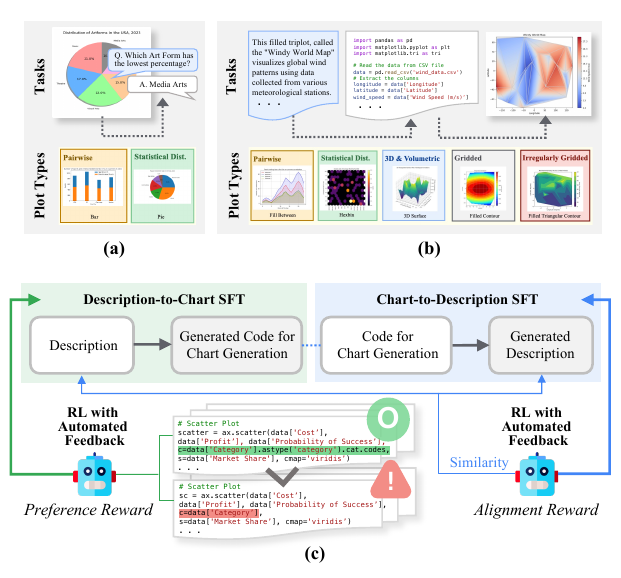}
    \caption{Illustration of the contributions of our method. (a): Existing datasets rarely cover a full range of chart types and primarily focus on QA tasks rather than chart generation. (b): Our dataset focuses on chart generation tasks and covers 31 unique plot types with tuples that combine descriptions, code, data tables, intermediate reasoning steps, and plots. (c): We further adopt RL-based instruction tuning method that leverage automated feedback and cycle consistency.}
    \label{fig:main}
\end{figure}

Recently, a range of NLP tasks has been addressed by leveraging the remarkable ability of Large Language Models (LLMs). This advancement has been possible largely through the process of instruction-tuning \citep{ouyang2022training,yoo2024hyperclova}, which fine-tunes LLMs to rely on intuitive natural language instructions and skillfully solve intricate tasks, encompassing fields like question answering \citep{sanh2022multitask,liu2023goat}, summarizing \citep{goyal2023news,fetahu2023instructpts}, and sentiment analysis \citep{varia2023instruction}. However, available LLMs continue to suffer from the difficult tasks of visualizing complex, fact-based, real-world data through charts and plots, mainly because of two challenges.

Firstly, the current datasets \citep{methani2020plotqa,masry2022chartqa,kahou2018figureqa,zhu2021autochart,kantharaj2022charttotext,han2023chartllama} primarily focus on QA in the chart domain rather than chart generation, and they rarely cover a full range of chart types and their varied applications. Several chart forms like 3D, volumetric, gridded, and irregularly gridded remain largely unexplored or insufficiently studied. These forms are important for evaluating the capabilities of LLMs in understanding multidimensional data, spatial data, and vector field data. Developing such instructional datasets typically entails significant expenses due to the complex nature of text-to-chart processes, incorporating various data components such as text, code, and data tables. 
This complexity, along with the lack of specific online sources containing these plot types, makes their collection difficult and time-consuming. It necessitates human expert intervention to ensure quality, which drives up costs.

Secondly, existing instruction-tuning methods based on supervised fine-tuning do not fully utilize the potential of rich datasets; for example, chart data include multiple components like text descriptions, code, and figures. 
Supervised fine-tuning struggles to effectively extract and leverage all the intricate information and relationships within these components, leading to suboptimal performance.


To address the first challenge, we propose a novel hierarchical pipeline for chart generation by leveraging the advanced linguistic skills of GPT-3.5-turbo \citep{ouyang2022training} and code generation and data analysis capabilities of GPT-4-0613 \citep{openai2023gpt4}. We contribute a dataset encompassing 31 unique plot types from the Matplotlib library \citep{Hunter:2007}, featuring 11.1K tuples that combine descriptions, code, data tables, and plots, covering a wide range of use cases.
Our pipeline is structured into the following steps: topic generation, description creation, code production, data table and reasoning step formulation, and cycle consistency verification.
This approach reduces biases towards common topics or plot types, and ensures consistent and accurate generation of multiple data elements.
By minimizing the human supervision in our proposed pipeline, we can generate a high-quality large-scale dataset that includes comprehensive descriptions, codes, data tables, reasoning steps, and illustrated graphs.


We further propose a novel reinforcement learning-based instruction tuning technique to address the second challenge. This method is tailored for chart generation tasks without costly human feedback.
We propose two different reward functions: the preference reward and alignment reward. 
For the preference reward, we construct a preference dataset from the supervised fine-tuned model's output and the ground truth code.
For the alignment reward, we optimize the model to increase the similarity between ground truth description and regenerated description from the code, exploiting the cycle consistency between code and description.
We jointly optimize two sequential policy models using the PPO \citep{schulman2017proximal}.


Finally, we make the following contributions:
\begin{itemize}[leftmargin=7.5mm]
\setlength{\itemsep}{2pt}
    \item We develop a novel dataset generation pipeline that populates data samples and filters out the low-quality ones, exploiting the cycle consistency in the task. 
    This approach is scalable to increase the volume of data as needed.
    \item We introduce the Text2Chart31 dataset, comprising 31 plot types with 11.1K tuples that combine descriptions, code, data tables, intermediate reasoning steps, and plots, covering a wide range of use cases.
    \item 
    We introduce an RL-based instruction tuning method that utilizes novel reward functions that leverage automated feedback and cycle consistency. The experiments demonstrate that our fine-tuned models outperform state-of-the-art open and closed-source models on data visualization tasks. To the best of our knowledge, this is the first work to adopt an RL-based instruction tuning approach for the chart generation task.
\end{itemize}

\section{Text2Chart31 Dataset}
Our newly contributed Text2Chart31 dataset supports 31 plot types based on Matplotlib
with 11.1K data points. 
We outline its key characteristics in \Cref{tab:comparison} comparing with existing datasets in the data visualization domain. 
The Text2Chart31 dataset $\mathcal{D}$ consists of 11,128 data points, each of which contains a tuple of ${(x, c, d, r, y)}$: a textual plot description ($x$), its corresponding code ($c$), the resulting plots ($y$).
For 8,166 data points, we additionally include a raw data table ($d$) and intermediate reasoning steps ($r$) to generate descriptions. 

\begin{figure*}[t]
    \centering
    \includegraphics[width=0.9\textwidth]{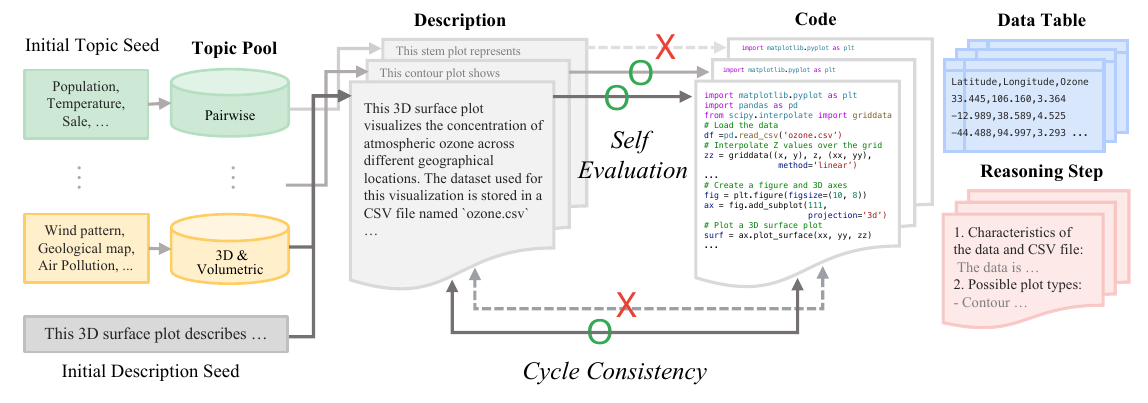}
    \caption{Illustration of our hierarchical chart generation process with an example of a single plot type. The process begins by randomly selecting a topic from a topic pool. Two instructional samples are then chosen from an instruction pool and given to GPT-3.5-turbo to generate a new instruction, which undergoes a self-evaluation process by GPT-4 for qualification. If it meets the criteria, which includes compatibility with the data points and the plot type, it is added to the instruction pool. Simultaneously, the new instruction is sent to GPT-4 for data table creation using a long data table format and code generation. Finally, the generated tuple ($x, d, c, y$) goes through a final filtering of cycle-consistency to validate the produced data point with high quality and correctness.}
    \label{fig:pipeline}
\end{figure*}

For the dataset, we develop a hierarchical plot generation pipeline leveraging GPT-3.5-turbo and GPT-4. 
Despite their impressive capabilities for text and code generation, collecting high-quality data points is challenging for two primary reasons: (1) GPT-3.5-turbo exhibits bias towards particular topics or narrow plot types that are commonly represented in its training data, and (2) the text-to-chart data involves multiple data elements including descriptions, code, and data tables, making it difficult to generate accurate and consistent data points in a single step. Consequently, we claim that a hierarchical approach is essential for producing higher-quality chart-generation data points. This pipeline is illustrated in \Cref{fig:pipeline}.

\begin{table*}[t]

\centering
\begin{adjustbox}{max width=2.0\columnwidth}
\begin{tabular}{l  c c c c c c c  c  c  c c}
\toprule
& \multicolumn{4}{c}{\small{\# Data }} 
& \multicolumn{4}{c}{\small{\# Plot Type }}
& \multicolumn{2}{c}{\small{Quality Analysis}} \\
\cmidrule(lr){1-1}
\cmidrule(lr){2-5}
\cmidrule(lr){6-9} 
\cmidrule(lr){10-11}

\small{{Dataset}}&\scriptsize{Figures} & \shortstack{\scriptsize{Instruction}\\\scriptsize{Tuning}}& \shortstack{\scriptsize{Description} \\ \scriptsize{to Code}}  & \shortstack{\scriptsize{Raw Data} \\ \scriptsize{to Description}} & 
\shortstack{\scriptsize{Pairwise\&} \\ \scriptsize{Stat. Dist.}} & 
\shortstack{\scriptsize{(Irregularly}) \\ \scriptsize{Gridded}}  &
\shortstack{\scriptsize{3D \&} \\ \scriptsize{Volumetric}}
&\scriptsize{Total}& \shortstack{ \scriptsize{Dataset} \\ \scriptsize{Balance$^\dagger$}  } &  \shortstack{\scriptsize{Content}\\ \scriptsize{Diversity}$^\ddagger$}\\  

\cmidrule(lr){1-1}
\cmidrule(lr){2-5}
\cmidrule(lr){6-9} 
\cmidrule(lr){10-11}

\small{PlotQA} & \small{224.3K} &\small{28.9M} & \small{\xmark} & \small{\xmark} & \small{3} & \small{\xmark}& \small{\xmark} &\small{3} & \small{0.786} &  \small{0.038} \\

\small{ChartQA} & \small{21.9K} & \small{32.7K} & \small{\xmark} & \small{\xmark} & \small{3} & \small{\xmark}& \small{\xmark} &\small{2} & \small{0.422} &  \small{-} \\

\small{FigureQA} & \small{180K} & \small{2.3M} & \small{\xmark}& \small{\xmark} & \small{4} & \small{\xmark} & \small{\xmark} & \small{4} & \small{0.960} &  \small{-} \\

\small{Unichart} & \small{611K} & \small{7M} & \small{\xmark} & \small{\xmark} & \small{3} & \small{\xmark} &\small{\xmark} & \small{3} & \small{0.821} &  \small{0.157}\\

\small{AutoChart} & \small{10.2K} & \small{23.5K} & \small{\xmark} & \small{\xmark} & \small{3} & \small{\xmark} &\small{\xmark} & \small{3} & \small{0.978} & \small{0.027}\\

\small{Chart-to-Text} & \small{44K} & \small{44K}& \small{\xmark} & \small{\xmark} &  \small{5} & \small{1} & \small{\xmark} & \small{6} & \small{0.327} & \small{0.421}\\
\small{ChartLlama} & \small{11K} & \small{160K} &  \small{7.8K} & \small{\xmark}  & \small{8}  & \small{2} & \small{\xmark} & \small{10} & \small{0.738} & \small{-}\\

\small{ChartX} & \small{6K} & \small{48K} &  \small{6K} & \small{\xmark}  & \small{13}  & \small{2} & \small{1} & \small{16} & \small{0.953} & \small{0.534}\\
\cmidrule(lr){1-1}
\cmidrule(lr){2-5}
\cmidrule(lr){6-9} 
\cmidrule(lr){10-11}
\small{Text2Chart31}& \small{11.1K} & \small{19.3K} & \small{\textbf{11.1K}} & \small{\textbf{8.2K}}  & \small{\textbf{16}}  & \small{\textbf{10}} & \small{\textbf{5}} & \small{\textbf{31}} & \small{0.980} & \small{0.674} \\
 \small{Text2Chart31-v2$^\S$}& \small{28.2K} & \small{50.2K} & \small{\textbf{28.2K}} & \small{\textbf{22K}}  & \small{\textbf{16}}  & \small{\textbf{10}} & \small{\textbf{5}} & \small{\textbf{31}} & \small{\textbf{0.993}} & \small{\textbf{0.696}} \\
\bottomrule
\end{tabular}
\end{adjustbox}
\caption{
\looseness=-1 Comparison with other chart datasets: PlotQA \citep{methani2020plotqa}, ChartQA \cite{masry2022chartqa}, FigureQA \cite{kahou2018figureqa},
Unichart \cite{masry2023unichart}, Autochart \cite{zhu2021autochart}, Chart-to-Text \cite{kantharaj2022charttotext}, ChartLlama \cite{han2023chartllama}, and ChartX \cite{xia2024chartx}. 
We report the total number of figures and instruction tuning data, including the tasks like QA, summarization, code generation, and plot recommendation. Additionally, we provide the number of data points for the tasks of \textit{Description to Chart} and \textit{Raw Data to Chart}, specifying data for Description to Code (visualization code) and Raw Data Analysis to Description (analyzing raw data to generate a corresponding description).
We also detail the number of plot types in each dataset.
$^\dagger$We measure the dataset balance score using the Shannon Diversity Index \citep{friedman2023vendi}.
$^\ddagger$We evaluate the content diversity by calculating average distinct $n$-grams ($n$ from 1 to 5) \citep{li2016diversitypromoting}. For PlotQA, Chart-to-Text, and AutoChart, we use chart titles, captions, and descriptions to evaluate content diversity, respectively. For Unichart and ChartX, we use summarizations. ChartQA and FigureQA are excluded due to lack of descriptions/titles, and ChartLlama is private. Finally, content diversity of Text2Chart31 is computed using the topics.
$^\S$ Text2Chart31-v2 is constructed and published at the camera ready version of the paper, and the experiment results in this paper is conducted with Text2Chart31.
}

\label{tab:comparison}
\end{table*}

\subsection{Task Definition}

Our benchmark is designed to evaluate three tasks. (1) \textit{Description-to-Chart:} Given a plot description $x$, an algorithm generates its corresponding code $c$ that creates a chart by the Matplotlib library\footnotemark 
 \citep{Hunter:2007}. \footnotetext{We use Matplotlib 3.8 version.}(2) \textit{Raw Data-to-Chart:} When provided with only a raw data table $d$, the algorithm generates intermediate reasoning steps $r$ that analyze the raw data and then generates a description $d$ for the most suitable plot type based on the characteristics of the data. 
(3) \textit{Code-to-Description:} Given the code $c$ for a plot, the model generates a detailed description $x$ of the plot.

\subsection{Dataset Construction Pipeline}
Our pipeline initiates by generating a topic from which a description $x$ is derived. To ensure both diversity of the topic and alignment with the intended plot type, each topic is filtered before proceeding to the next step. We additionally generate code $c$, raw data table $d$ and intermediate reasoning step $r$ corresponding to the description. Lastly, we use the cycle-consistency verification to ensure the high quality of the data points.
Please refer to \Cref{detail:cycle} for the detailed process with examples.

\textbf{Topic generation}.
We generate distinct topic pools for five different plot categories: pairwise, statistical, gridded, irregularly gridded, and 3D/volumetric data. To maintain diversity within each topic pool, we include only topics with low similarity scores compared to those already being presented. To assess similarity, the ROUGE-L metric \citep{lin-2004-rouge} is employed as a common practice from previous studies \citep{wang2023selfinstruct}. 

\textbf{Description generation}.
For each plot type, we start by manually writing 5 to 10 descriptions as seed points that contain all the necessary information for a plot to be illustrated.  To generate a description ($x$), we randomly sample two descriptions and pair them with a topic from the topic pool. This assembled data is prompted into GPT-3.5-turbo, which generates a similar format plot description for the sampled topic. We remove the topic from the pool after a new description is generated to uphold the diversity. Inspired by the studies on the reasoning capabilities of LLMs \cite{wei2023chainofthought,kojima2023large,wang2023selfconsistency}, we instruct GPT-4 to \textit{self-evaluate} the generated descriptions for quality control. This step is crucial to exclude any incompatible instructions that can lead to the creation of unsuitable plots, thereby avoiding computational waste.

\textbf{Code generation}.
We input descriptions into GPT-4, which is instructed to generate Python code for the Matplotlib library. This code aims to  visualize the described plot. We add the generated code ($c$) to the dataset only if it successfully generates the corresponding plot ($y$) without a runtime error.

\textbf{Data table and reasoning step generation}.
For plots derived from data files in $\mathcal{D}$, GPT-4 is prompted to generate either a raw data table $d$ or Python code that can generate the data table. 
3D volumetric, gridded, and irregularly gridded plots often require specific patterns or mathematical relations between variables; therefore, code is created and executed to generate the data table instead of directly generating it. 
We further generate intermediate reasoning steps $r$ using GPT-4, which is instructed to analyze the characteristics of the data and CSV file, explore possible plot types, determine the most suitable plot type, and consider additional aspects of the description. This process results in data points $(x,c,d,r,y)$. 



\textbf{Cycle-Consistency verification}.
We argue that given the complex and fact-based nature of text-to-chart datasets, employing human evaluation to check the quality of generated data points is inefficient. To this end, we propose an AI-assisted method using cycle consistency, to assure the quality of the data point. This process involves regenerating an instruction that describes the plot from the generated code and comparing it against the original one. We keep the data only if the regenerated description closely aligns with the original one based on pre-defined criteria,  indicating the high quality of the data. We provide further details on the cycle consistency method in \Cref{app:prompt}.

\subsection{Analysis of Text2Chart31 Dataset}


As shown in Table \ref{tab:comparison}, 
we can effectively balance the data points per plot type with equal distribution in the dataset, which is quantified by the Shanon Diversity metric \citep{friedman2023vendi}. 
Shannon Diversity is computed through $H=-\sum_{i=1}^S p_i \log(p_i)$, where $S$ is the total number of classes in the dataset, and $p_i$ is the proportion of instances belonging to the $i$-th class. 
Our Text2Chart31 dataset achieve the highest score of 0.981.
\Cref{fig:chart-type-dist} in Appendix shows a detailed comparison of the distribution per chart type between datasets using pie charts.
We further evaluate the content diversity of datasets via  Distinct-n score \citep{li2016diversitypromoting}. Our dataset achieves a score of 0.674, indicating that our pipeline effectively reassures the diversity of topics.

\begin{algorithm*}[t]
\caption{Chart generation instruction tuning}
\begin{algorithmic}[1]
\Require Description-to-chart policy network $\pi_{\theta_{1}}$, 
Raw data-to-chart policy network $\pi_{\theta_{2}}$,
code-to-description policy network $\pi_{\theta_{3}}$, Text2Chart31 dataset $\mathcal{D}$
\For{iter = $1, 2, \ldots, N_\mathrm{sft}$} \Comment{Supervised fine-tuning}
\State Sample data $(x, c, d, r, y)$ from dataset $\mathcal{D}$
\State Optimize $\mathcal{L}_\mathrm{code}(\theta_1) = - \sum_{t} \log \pi_{\theta_1} (c_{(t)}| x, c_{(<t)})$ \label{line:sft1}
\State Optimize $\mathcal{L}_\mathrm{reason}(\theta_2) + \mathcal{L}_\mathrm{desc}(\theta_2) = - \sum_{t} \log \pi_{\theta_2} (r_{(t)}| d, r_{(<t)}) - \sum_{t} \log \pi_{\theta_2} (x_{(t)}| d,  r, x_{(<t)})$ \label{line:sft2}
\State Optimize $\mathcal{L}_\mathrm{desc}(\theta_3) = - \sum_{t} \log \pi_{\theta_3} (x_{(t)} |x_{(<t)}, c, d)$ \label{line:sft3}
\EndFor
\State $\pi_{\mathrm{\theta}_\mathrm{sft1}} \leftarrow \pi_{\theta_1}$, 
$\pi_{\mathrm{\theta}_\mathrm{sft2}} \leftarrow \pi_{\theta_2}$, 
$\pi_{\mathrm{\theta}_\mathrm{sft3}} \leftarrow \pi_{\theta_3}$
\State Generate automatic preference dataset $\mathcal{D}_\mathrm{pref}$ from $\pi_{\mathrm{\theta}_\mathrm{sft1}}$ and $\mathcal{D}$
\State Train preference reward model $R_\phi(c)$ from $\mathcal{D}_\mathrm{pref}$
\For{iter = $1, 2, \ldots, N_\mathrm{rl}$}  \Comment{Reinforcement learning (PPO)}
\State Sample $x$ from dataset $\mathcal{D}$
\State Generate $\hat{c}$ from $\pi_{\mathrm{\theta}_1}(\cdot| x)$, and generate $\hat{x}$ from $\pi_{\mathrm{\theta}_3}(\cdot| \hat{c})$
\State Calculate preference reward $R_\phi(\hat{c})$
\State Calculate alignment reward $R(x, \hat{x}) = \mathrm{BertScore}(x, \hat{x})$
\State Jointly optimize $\Bigg(J_\mathrm{PPO}(\theta_1) = R_\phi(\hat{c})  - \beta \log \left(\frac{\pi_{\theta_1}(\hat{c} \,|\, x)}{\pi_{\mathrm{\theta}_\mathrm{sft1}}(\hat{c} \,|\, x)}\right)$,

~~~~~~~~~~~~~~~~~~~~~~~~~~~$J_\mathrm{PPO}(\theta_3) = R(x, \hat{x}) - \beta \log \left(\frac{\pi_{\theta_3} (\hat{x}\,|\, \hat{c})}{\pi_{\mathrm{\theta}_{\mathrm{sft3}}} (\hat{x} \,|\, \hat{c})}\right)\Bigg)$ with PPO
\EndFor
\end{algorithmic}
\label{alg:chart}
\end{algorithm*}

\section{Instruction Tuning Approach}

We discuss our proposed instruction tuning methods for fine-tuning LLMs to tackle the three data visualization tasks: (1) Description-to-Chart, (2) Raw-Data-to-Chart, and (3) Code-to-Description, using the Text2Chart31 dataset.
We respectively denote three specialized models for the three tasks: $\pi_{\theta_1}$, $\pi_{\theta_2}$, and $\pi_{\theta_3}$.
We train these models with two phases: supervised fine-tuning (SFT), followed by reinforcement learning (RL) with two types of reward that are specifically tailored to improve chart generation performance. Initially, all three tasks undergo supervised fine-tuning.
Afterward, using PPO algorithm \citep{schulman2017proximal},  we jointly optimize $\pi_{\theta_1}$ with the preference reward and $\pi_{\theta_3}$ with the alignment reward that ensures cycle consistency and coherence of outputs. 
\Cref{alg:chart} summarizes the overall procedure.

\subsection{Supervised Fine-tuning}

We perform supervised fine-tuning of $\pi_{\theta_1}$, $\pi_{\theta_2}$, and $\pi_{\theta_3}$ using the cross-entropy loss with the Text2Chart31 dataset. 
For Task 1, the model $\pi_{\theta_1}$ maximizes the probability of outputting the ground truth code for a given description by minimizing cross-entropy loss in the Line \ref{line:sft1} of \Cref{alg:chart}.
For Task 2, we design the model $\pi_{\theta_2}$ to generate descriptions from raw data in two stages. 
First, the model generates a reasoning step $r$ from the raw data $d$, which involves analyzing data characteristics and determining the appropriate plot type.
Then, the model is fine-tuned to generate the description $x$ using the data and the reasoning step as in the Line \ref{line:sft2}.
Lastly, we fine-tune the model $\pi_{\theta_3}$ for Task 3 to maximize the probability of predicting the ground truth description for a given visualization code as in the Line \ref{line:sft3} of \Cref{alg:chart}.

\subsection{RL via Automatic Feedback} 

We design two reward functions, which are the preference reward and the alignment reward, specifically tailored for the chart generation task.
It is worth noting that we remove human supervision during these processes and solely rely on automatic feedback.

\textbf{Preference reward}. 
We propose an automatic way of designing a preference dataset based on the output of the supervised fine-tuned model $\pi_{\theta_1}$.
We define preference dataset $\mathcal{D}_\mathrm{pref} = (c^+_i, c^-_i)_{i=1}^n$, where a preferred code $c^+$ is the ground truth code, while a less preferred one $c^-$ is a corresponding code output of SFT.
Afterward, we train a preference reward model $R_\phi(c)$ following \citet{ouyang2022training} and employ this reward model to train $\pi_{\theta_1}$ via proximal policy optimization (PPO) algorithm \citep{schulman2017proximal} as follows:
\begin{align*}
\underset{\theta_1}{\mathrm{maximize}}\; &\mathbb{E}_{
\substack{x \sim \mathcal{D},\,
\hat{c} \sim \pi_{\theta_1}(\cdot | x)}} \biggr(R_\phi(\hat{c}) \biggl) \\
&- \beta\infdiv{\pi_{\theta_1}}{\pi_{\theta_\mathrm{sft1}}}. 
\end{align*}
 
\textbf{Alignment reward.} 
The alignment reward leverages cycle consistency between a chart's description and  code. 
First, $\pi_{\theta_1}$ generates a code from the original description, then $\pi_{\theta_3}$ uses this code to produce a regenerated description. The alignment reward is defined as the similarity between the original and regenerated descriptions, measured by BertScore \citep{zhang2020bertscore, black2024training}.
We optimize $\pi_{\theta_3}$ via maximizing the alignment reward $R(\cdot, \cdot)$ using PPO algorithm as follows:
\begin{align*}
\underset{\theta_3}{\mathrm{maximize}}\; &\mathbb{E}_{
\substack{x \sim \mathcal{D},\,
\hat{c} \sim \pi_{\theta_1}(\cdot | x), \, 
\hat{x} \sim \pi_{\theta_3}(\cdot | \hat{c})}
} \biggr(R(x, \hat{x}) \biggl) \\
&- \beta\infdiv{\pi_{\theta_3}}{\pi_{\theta_\mathrm{sft3}}}. 
\end{align*}

\section{Experiments}

\textbf{Baselines.}
For the evaluation of the three target tasks, we compare with the state-of-the-art open-source baseline models as follows: (i) Description-to-Chart: Code Llama Instruct \citep{roziere2024code}, Llama 3 Instruct \citep{llama3}, StarCoder \citep{li2023starcoder}, and Instruct CodeGen \citep{nijkamp2023codegen}, (ii) Raw Data-to-Description: Llama 2 Chat \citep{touvron2023llama} and Llama 3 Instruct model, and (iii) Code-to-Description: Code Llama, Llama 2 Chat, and Llama 3 Instruct models.
We also compare with proprietary models including GPT-3.5-turbo \citep{ouyang2022training}, GPT-4-0613, GPT-4-turbo-2024-04-09 \citep{openai2023gpt4}, GPT-4o-2024-05-13 \citep{gpt-4o}, and Claude 3 Opus \citep{claude3}.

\textbf{Evaluation metrics.}
For the three target tasks, we report the following evaluation measures. 

(i) Description-to-Chart: We report the total error ratio and plot-type error ratio. The total error ratio indicates the percentage of code executions that result in errors. We categorize and report plot-type errors based on Matplotlib classifications. 
We further evaluate the similarity between the predicted code and the ground truth (GT) code by reporting the METEOR \citep{banerjee-lavie-2005-meteor} and CodeBLEU metrics \citep{ren2020codebleu}.

(ii) Raw Data-to-Description: 
We report the Jaccard similarity and the Hit Rate. The former measures the intersection ratio between the recommended plot list derived from generated reasoning steps and the GT reasoning steps. The latter is the percentage of recommended lists containing the GT plot type.
To evaluate the quality of the generated descriptions, we first use these descriptions to generate code with both the SFT Llama3 Instruct-8B model and the GPT-3.5-turbo, and then calculate the error ratio for the generated codes. Additionally, we report ROUGE-L and BertScore metrics to assess the similarity between the generated descriptions and the GT descriptions. 

(iii) Code-to-Description: We measure ROUGE-1/2/L and BertScore to evaluate the similarity between the generated descriptions and the GTs. Lastly, as done for Task 2, we generate the code by giving the predicted descriptions to the GPT-3.5-turbo and report the error ratio.  
\begin{table*}[t]

\centering
\begin{adjustbox}{max width=2.0\columnwidth}
\begin{tabular}{lccccccc}
\toprule
& \multicolumn{5}{c}{Error ratio (\%) $\downarrow$} & \multicolumn{2}{c}{Code similarity $\uparrow$}
\\
\cmidrule(r){2-6} \cmidrule(r){7-8}
& & Statistical &  (Irregularly) & 3D \&  &  & & \\
Model & Pairwise & distribution    
& gridded & Volumetric  & Total  & METEOR & CodeBLEU \\
\cmidrule(r){1-1}\cmidrule(r){2-5} \cmidrule(r){6-6} \cmidrule(r){7-8}
\multicolumn{3}{l}{\textit{\textbf{Open-source}}} &  &  &  &  &  \\
\cmidrule(r){1-1}\cmidrule(r){2-5} \cmidrule(r){6-6} \cmidrule(r){7-8}

    CLI-7B  & 22.67 & 29.42 & 77.94 & 52.20 & 41.32 & 0.485 & 0.402 \\
    L3I-8B  & 20.76 & 28.98 & 66.76 & 34.59 & 35.91 & 0.519 &  0.437  \\
    \textbf{{[}SFT{]}} L3I-8B   & 19.07 & 13.27 & \textbf{13.53} & \textbf{20.75} & 16.09 & 0.562 &   \textbf{0.464} \\
    
    
    \textbf{{[}SFT+$\text{RL}_\text{pref}${]}} L3I-8B  & \textbf{13.14} & \textbf{11.50} & 15.00 & 26.42 & \textbf{14.55} & \textbf{0.567} & 0.461  \\
\cmidrule(r){1-1}\cmidrule(r){2-5} \cmidrule(r){6-6} \cmidrule(r){7-8}
    CLI-13B   &18.86 & 29.42 & 71.76 & 57.23 & 39.14 & 0.489 &  0.413  \\
    StarCoder-15.5B  & 23.31 & 32.08 & 51.18 & 25.16 & 32.89 & 0.347 & 0.328 \\
    Instruct CodeGen-16B & 38.56 & 45.13 & 62.94 & 40.25 & 46.66 & 0.388 &  0.330  \\
    \textbf{{[}SFT{]}} CLI-13B   & \textbf{6.36} & 6.19 & \textbf{12.06} & 22.64 &9.49 & \textbf{0.581} & \textbf{0.481}   \\
     \textbf{{[}SFT+$\text{RL}_\text{pref}${]}} CLI-13B   & \textbf{6.36} & \textbf{5.53} & 12.35 & \textbf{21.38} &\textbf{9.21} & 0.566 &  0.467  \\
\cmidrule(r){1-1}\cmidrule(r){2-5} \cmidrule(r){6-6} \cmidrule(r){7-8}
\multicolumn{3}{l}{\textit{\textbf{Closed-source}}} &  &  &  &  &  \\
\cmidrule(r){1-1}\cmidrule(r){2-5} \cmidrule(r){6-6} \cmidrule(r){7-8}
    GPT-3.5-turbo & 11.02 & 13.50 & 28.82 & 19.59 & 18.62 & 0.524 & 0.453\\
    
    GPT-4-0613 & 13.56 & 11.06 & 28.53 & 39.62 & 19.26 & 0.535 & 0.441 \\
    GPT-4-turbo & 11.02 & 14.16 & 11.76 & 29.56 & 14.27 & 0.540 & 0.448 \\ 
    GPT-4o & 13.98 & 6.86 & 13.53 & 26.42 & 13.00 & 0.552& 0.450\\
    Claude 3 Opus &7.84 & 7.74 & 30.59 & 23.27 & 14.90 &0.515 & 0.435 \\
\bottomrule
\end{tabular}
\end{adjustbox}

 \caption{Results of the Description-to-Chart task. 
The plot type error ratio is categorized based on Matplotlib classifications \citep{Hunter:2007}. CLI and L3I stand for Code Llama Instruct and Llama 3 Instruct, respectively. \textbf{SFT} and  \textbf{$\text{RL}_\ast$} indicate our fine-tuned models.}
\label{tab:task1}
\end{table*}


\textbf{Training setup.}
We begin the supervised fine-tuning using LoRA fine-tuning \citep{hu2021lora}. 
When we further fine-tune the model with RL, we merge the original SFT LoRA parameters into the base model and fine-tune separate LoRA parameters. 
For SFT, we utilize a total of 11.1K data points for Task 1, 3, and 7.84K for Task 2. On the other hand, RL fine-tuning is conducted using 0.5K randomly selected data points, representing 4.8\% of our $\mathcal{D}_\mathrm{pref}$ dataset. 
For SFT, we use 2 RTX A6000 GPUs and the training requires 6 to 12 hours, depending on the tasks. For RL, we use 6 RTX A6000 GPUs and the training takes less than 12 hours.
Further details of the experiments can be found in \Cref{app:detail}.

\begin{table}[t]

\centering
\begin{adjustbox}{max width=1.0\columnwidth}
\begin{tabular}{lccccccc}
\toprule 
& \multicolumn{2}{c} { Error ratio (\%) $\downarrow$} & \multicolumn{2}{c}{Plot type  $\uparrow$}  &\multicolumn{2}{c} { Desc. sim. $\uparrow$} \\
\cmidrule(r){2-3} \cmidrule(r){4-5} \cmidrule(r){6-7} 
Method & w/ GPT & w/ SFT & HitRate  & Jac.  & R-L & BertScore   \\
\cmidrule(r){1-1}\cmidrule(r){2-3} \cmidrule(r){4-5} \cmidrule(r){6-7} 
\multicolumn{3}{l}{\textit{\textbf{Open-source}}} &  &  \\
\cmidrule(r){1-1}\cmidrule(r){2-3} \cmidrule(r){4-5} \cmidrule(r){6-7} 
    L2C-7B &40.82  & 56.45	 &0.175	 &0.359	 &0.232	 &0.820\\
    L2C-13B 	&35.64  &37.99 & 0.205 &0.384	 &0.237	 &0.825\\
    L3I-8B  &27.25	&38.48	&0.269	&0.406	&0.196	&0.800\\
    \textbf{{[}SFT{]}} 
    L2C-7B & 15.92	&15.82	&0.329	&0.396	&0.381	&0.903\\
    \textbf{{[}SFT{]}} L3I-8B & \textbf{15.53} & \textbf{15.62}	 &\textbf{0.413}& \textbf{0.432}	&\textbf{0.389} &\textbf{0.905}\\
\cmidrule(r){1-1}\cmidrule(r){2-3} \cmidrule(r){4-5} \cmidrule(r){6-7} 
\multicolumn{3}{l}{\textit{\textbf{Closed-source}}} \\
\cmidrule(r){1-1}\cmidrule(r){2-3} \cmidrule(r){4-5} \cmidrule(r){6-7} 
    GPT-3.5-turbo &21.00& 29.10	&0.239	 &0.412 &0.232	 &0.816\\
    GPT-4	&16.41	&34.67&	0.286	&0.428	&0.202	&0.829 \\
    GPT-4-turbo & 27.05	&37.01	&0.313	& 0.461	& 0.184	&0.808 \\
    GPT-4o & 15.82	&31.64&	0.339	&0.436 &	0.170&	0.786\\
    Claude 3 Opus &15.62	&27.34  &0.294	&0.451	&0.188	&0.813 \\
\bottomrule
\end{tabular}
\end{adjustbox}

\caption{Results of the Raw Data-to-Chart task. Description similarity, error ratio, and plot type prediction are compared for various open-source and closed-source methods. 
The error ratio is evaluated using \textbf{SFT} L3I-8B from Task 1 denoted as 'w/ SFT', or GPT-3.5-turbo denoted as 'w/ GPT'. 
\textbf{SFT} indicates our fine-tuned models. L2C and L3I stand for Llama 2 Chat and Llama 3 Instruct, respectively.}
\label{tab:task2}
\end{table}

\begin{table}[t]

\centering
\begin{adjustbox}{max width=1.0\columnwidth}
\begin{tabular}{lccccc}
\toprule 
& \multicolumn{4}{c}{Description similarity $\uparrow$} & Err. ratio (\%) $\downarrow$ \\
\cmidrule(r){2-5}
\cmidrule(r){6-6}
Method & R-1  & R-2 & R-L & BertScore & w/ GPT \\
 
\cmidrule(r){1-1}\cmidrule(r){2-6} 
\multicolumn{3}{l}{\textit{\textbf{Open-source}}} &  &  \\
\cmidrule(r){1-1}\cmidrule(r){2-6} 
     L2C-7B & 0.419 & 0.182 & 0.260 & 0.812 & 38.86 \\
     CLI-7B  & 0.411 & 0.173 & 0.260 & 0.823  & 42.73 \\
    L3I-8B  & 0.453 & 0.206 & 0.276 & 0.834 & 36.82 \\
    \textbf{{[}SFT{]}} 
    L3I-8B  & 0.592  & 0.343 & 0.436 & 0.881 & 21.36 \\
    \textbf{{[}SFT+$\text{RL}_\text{algn}${]}}
    L3I-8B & \textbf{0.594} & \textbf{0.346} & \textbf{0.440} & \textbf{0.884} & \textbf{20.31} \\
\cmidrule(r){1-1}\cmidrule(r){2-6} 
\multicolumn{3}{l}{\textit{\textbf{Closed-source}}} &  &  \\
\cmidrule(r){1-1}\cmidrule(r){2-6} 
    GPT-3.5-turbo & 0.463 & 0.212 & 0.286  & 0.845 & 45.26 \\
    GPT-4 & 0.426 & 0.175 & 0.252 & 0.809 & 23.12 \\
    GPT-4-turbo & 0.416 & 0.168 & 0.242 & 0.795 & 29.80 \\
    GPT-4o & 0.442 & 0.198 & 0.269 & 0.775 & 11.81 \\
     Claude 3 Opus & 0.453 & 0.207 & 0.276 & 0.827 & 19.33\\
     
\bottomrule
\end{tabular}
\end{adjustbox}
\caption{Results of the Code-to-Description task. 
\textbf{SFT} and  \textbf{$\text{RL}_\ast$} indicate our fine-tuned models. L2C and L3I stand for Llama 2 Chat and Llama 3 Instruct-8B, respectively.}
\label{tab:task3}
\end{table}

\subsection{Results of Description-to-Chart}

Table \ref{tab:task1} presents the results for the Description-to-Chart task.
We fine-tune Llama 3 Instruct-8B and Code Llama Instruct-13B on our Text2Chart31 dataset for five epochs. 
We run RL fine-tuning on the Llama 3 Instruct and Code Llama Instruct-13B using preference reward, denoted as $\mathrm{RL}_\mathrm{pref}$. 
The results show that our fine-tuned models outperform all open-source baselines that we compared. 
Specifically, the 13B model with SFT and RL achieves even a lower total error ratio than the state-of-the-art closed-source models like GPT-3.5-turbo, GPT-4, GPT-4-turbo, GPT-4o, and Claude 3 Opus. 
The RL fine-tuning reduces the total error ratio of the Llama 3 Instruct-8B model from 16.09 to 14.55, making it superior to the  Claude 3 Opus. 
Particularly, our models excel in generating underexplored plot types such as gridded, irregularly gridded, and 3D and volumetric plots, compared to open-source models.

\begin{figure}[t]
    \centering
    \includegraphics[width=1.0\columnwidth]{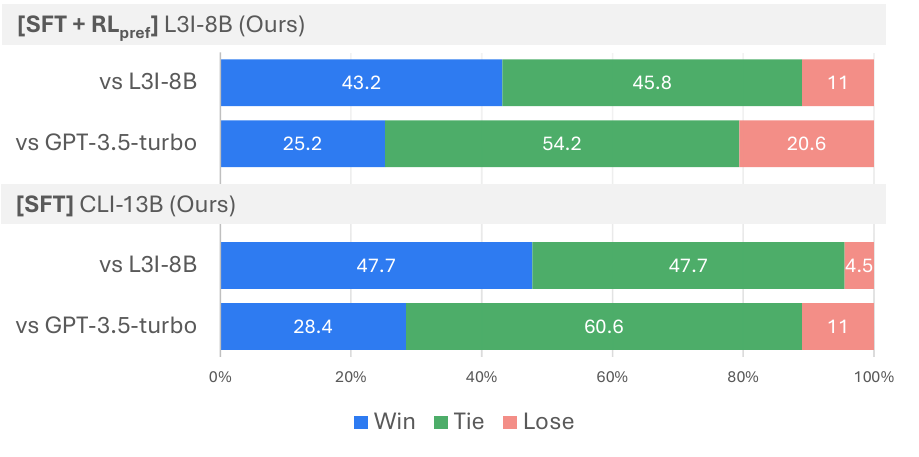}
    \caption{Human evaluation results on a randomly sampled subset of the test set. We compare \textbf{{}SFT+$\text{RL}_\text{pref}${}} L3I-8B and \textbf{{}SFT{}} CLI-13B with GPT-3.5-turbo and L3I-8B.}
    \label{fig:human-eval}
\end{figure}

\textbf{Human evaluation.}
We additionally conduct human evaluation to check the correctness of the generated plot and its alignment with the description. We randomly sample a subset of 155 data points, consisting of 5 samples from each of the 31 plot types. For each sample, three crowd workers are recruited to compare the generated plot images with the GT reference plot images based on chart type, data representation, and visual appearance. If both images are equally similar or neither is similar, it is voted as a tie. More details can be found in \Cref{app:human-eval}.
\Cref{fig:human-eval} presents the results of  human evaluation. The inter-annotator agreement is measured using Krippendorff's $\alpha$, whose value is 0.519 for the three classes (win, lose, and tie).
Our fine-tuned models consistently have higher win rate compared to Llama 3 Instruct-8B and GPT-3.5-turbo.
Specifically, \textbf{SFT} CLI-13B model has the higher win rate (47.7\%) against L3I-8B, while also achieving a lower lose rate (4.5\%). 
Our \textbf{{}SFT+$\text{RL}_\text{pref}${}} L3I-8B model wins over GPT-3.5-turbo with 25.2\% win rate and 20.6\% lose rate.

\subsection{Results of Raw Data-to-Chart}

Table \ref{tab:task2} presents the results of the Raw Data-to-Chart task. 
We fine-tune Llama 2 Chat-7B and Llama 3 Instruct-8B using our Text2Chart31 dataset. 
We report the error ratio after visualizing the generated descriptions using our supervised fine-tuned Llama 3 Instruct-8B (w/ SFT) from task 1 and GPT-3.5-turbo (w/ GPT). 
Notably, our fine-tuned Llama 3 Instruct-8B outperforms all open-source models across all metrics. 
Furthermore, this model surpasses closed-source models (GPT-3.5-turbo, GPT-4-turbo) in terms of error ratio and generated description similarity.

\subsection{Results of Code-to-Description}

Table \ref{tab:task3} presents the results on the Code-to-Description task. We fine-tune the Llama 3 Instruct-8B using our dataset and evaluate the description similarity with ROUGE  and BertScore. 
Our fine-tuned model outperforms all open-source and closed-source models across Description similarity.
Furthermore, RL fine-tuning with alignment reward consistently increases the description similarity across all metrics. We also provide the generated descriptions to GPT-3.5-turbo and report the error ratio to highlight the quality of the descriptions produced by our fine-tuned models. After RL fine-tuning, the error ratio decreases from 21.36\% to 20.31\%, and the description similarity consistently improves.  


\section{Related Work}
\textbf{Chart datasets.}
There are several existing chart datasets, including PlotQA \citep{methani2020plotqa}, ChartQA \cite{masry2022chartqa}, FigureQA \cite{kahou2018figureqa}, Unichart \cite{masry2023unichart}, Autochart \cite{zhu2021autochart}, Chart-to-Text \cite{kantharaj2022charttotext}. These datasets primarily focus on question and answer (QA) tasks on a limited range of plot types. More recently, ChartLlama \cite{han2023chartllama}  proposes a text-to-chart dataset that includes QA tasks and generates visualization code from provided descriptions. However, these datasets still lack coverage in certain plot categories such as 3D/volumetric plots and vector field plots, and they do not cover the use case of analyzing the raw data and predicting the most suitable plot types. On the other hand, our Text2Chart31 dataset encompasses 31 plot types with 11.1K tuples that combine descriptions, code, data tables, and plots, thereby covering a wide range of use cases.

\textbf{Instruction tuning}.
Employing reinforcement learning with human feedback is a prevalent strategy for enhancing (un)supervised finetuned models, whether by integrating human feedback into the learning loop \citep{arakawa2018dqntamer,arumugam2019deep} or by leveraging preference data generated by human \citep{ouyang2022training,glaese2022improving,bai2022training,stiennon2022learning}. However, we argue that this methodology might not offer the most practical solution for plot visualization tasks, given the intricate and fact-intensive nature of plot types. Moreover, considering the limitations of human cognition, there is a risk of overlooking crucial small details essential for validating the accuracy of generated plots. To address this, we propose a novel automatic method that constructs a preference dataset using supervised fine-tuned output. 

\textbf{Cycle consistency.}
Exploiting cycle consistency to enhance the performance of the generative model has been mainly studied in the image domain \citep{zhu2020unpaired}. Recently, DDPO \citep{black2024training} adopts the LLaVA model \citep{liu2023visual} to increase the alignment between the image and the text. Following this line of research, we propose an alignment reward that exploits cycle consistency between description and code to improve LLM for chart generation tasks. This is made possible because of the rich nature of our Text2Chart31 dataset, which consists of diverse textual modalities, including visualization code and description.

\section{Conclusion}

We introduce a novel hierarchical pipeline and a comprehensive dataset for chart generation. The proposed Text2Chart31 dataset, encompassing 31 unique plot types, provides a robust foundation for diverse visualization tasks with its 11.1K tuples of descriptions, code, data tables, and plots. 
Additionally, we proposed an RL-based instruction tuning technique employing preference and alignment rewards, 
improving LLMs in data visualization.

\section*{Limitations}
There are certain considerations to note. First, our dataset is based on Matplotlib version 3.8. As such, if earlier versions of Matplotlib are used where function names may have changed, the generated code could potentially cause errors. This is a natural consequence of advancements and updates in software libraries. Additionally, the descriptions provided are exclusively in English. This focus ensures clarity and consistency in our current scope but can be expanded to include multiple languages in future iterations. Lastly, our primary focus was on chart generation through large language models (LLMs), rather than on question answering. However, exploring question answering capabilities is a promising direction for future research.

\section*{Ethics Statement}

All data points generated in Text2Chart31 were created using large language models (LLMs) and are intended solely for visualization purposes. These data points do not represent real-world facts and should not be referenced as accurate depictions of actual data distributions.
Furthermore, they do not contain offensive contents.
Matplotlib library is based on PSF license. We have used open source models, libraries, and closed source models for their intended uses, and not use other than research purposes.

\section*{Acknowledgements}
We would like to thank the anonymous reviewers for their valuable feedback. This work was financially supported by SNU-NAVER Hyperscale AI Center, Institute of Information \& Communications Technology Planning \& Evaluation (IITP) grant funded by the Korea government (MSIT) (No.~RS-2019-II191082, SW StarLab, No.~RS-2022-II220156, Fundamental research on continual meta-learning for quality enhancement of casual videos and their 3D metaverse transformation, and No.~RS-2021-II211343, Artificial Intelligence Graduate School Program (Seoul National University)), and the National Research Foundation of Korea (NRF) grant funded by the Korea government (MSIT) (No.~2023R1A2C2005573).

\bibliography{custom}

\appendix
\clearpage
\onecolumn
\section{Details of \ours{} Dataset}
\label{app:text2chart31-details}
In this section, we provide a comprehensive overview of the \ours{} dataset, including its categories, examples, summary statistics, topic distribution and plot type distribution.

\subsection{Categories and Examples}
\label{app:category-exp}
\begin{figure}[ht]
    \centering
    \includegraphics[width=0.85\columnwidth]{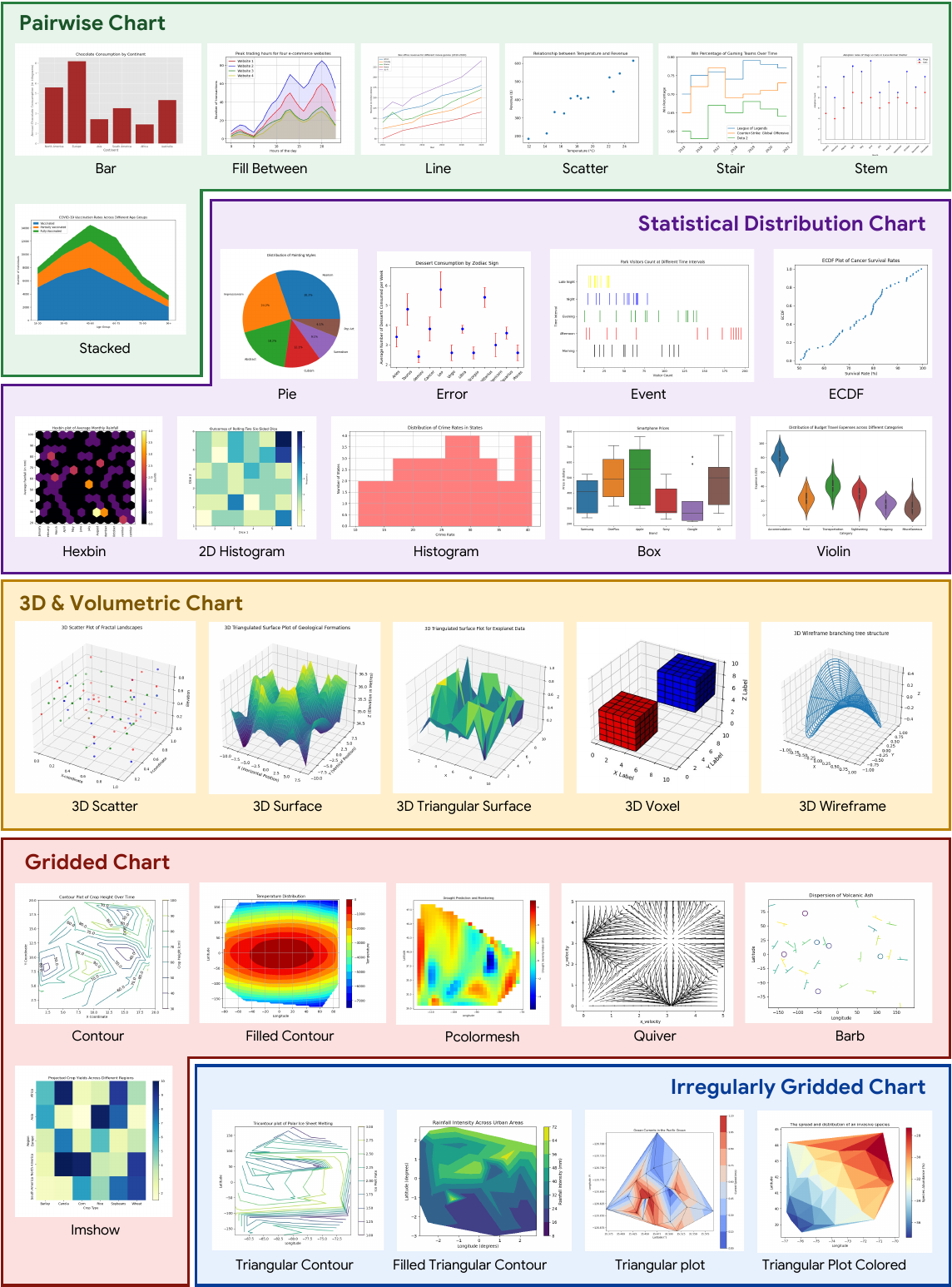}
    \caption{Examples from the 31 plot types in \ours{} dataset, grouped into 5 chart categories.}
    \label{fig:category-example}
\end{figure}

\Cref{fig:category-example} illustrates the diverse range of plot types included in the \ours{} dataset. The dataset covers 31 different plot types, grouped into 5 categories: Pairwise Chart, Statistical Distribution Chart, Gridded Chart, Irregularly Gridded Chart, and 3D \& Volumetric Chart. The examples provided for each plot type illustrate the variety of data and plot types present in the dataset.

\clearpage

\subsection{Dataset Summary}
\label{app:dataset-summary}
The \ours{} dataset consists of 11,128 data points, with 9,705 in the training set and 1,423 in the test set. The dataset is categorized into five categories of charts: Pairwise, Statistical Distribution, Gridded, Irregularly Gridded, and Statistical Distribution and 3D \& Volumetric chart. Among the total data points, 8,166 include both data tables ($d$) and reasoning steps ($r$), with 7,142 in the training set and 1,024 in the test set.

\begin{table}[ht]
\centering
\begin{tabular}{llll}
\toprule
                              & \textbf{Train} & \textbf{Test} & \textbf{Total} \\
\midrule
Pairwise Chart                 & 3026 (1557) & 472 (241)   & 3498 (1798)  \\
Statistical Distribution Chart & 2878 (1784) & 452 (284)   & 3330 (2068)  \\
Gridded Chart                  & 1305 (1305) & 192 (192)   & 1497 (1497)  \\
Irregularly Gridded Chart      & 1145 (1145) & 148 (148)   & 1293 (1293)  \\
3D \& Volumetric Chart         & 1351 (1351) & 159 (159)   & 1510 (1510)  \\
\midrule
\textbf{Total}                 & 9705 (7142) & 1423 (1024) & 11128 (8166) \\
\bottomrule
\\
\end{tabular}
\caption{Summary of the \ours{} dataset. The numbers in parentheses indicate the data points that include both data tables ($d$) and reasoning steps ($r$).}
\label{tab:dataset-summary}
\end{table}

\subsection{Topic Distribution}
\label{app:topic-dist}
\begin{figure}[ht]
    \centering
    \includegraphics[width=0.9\columnwidth]{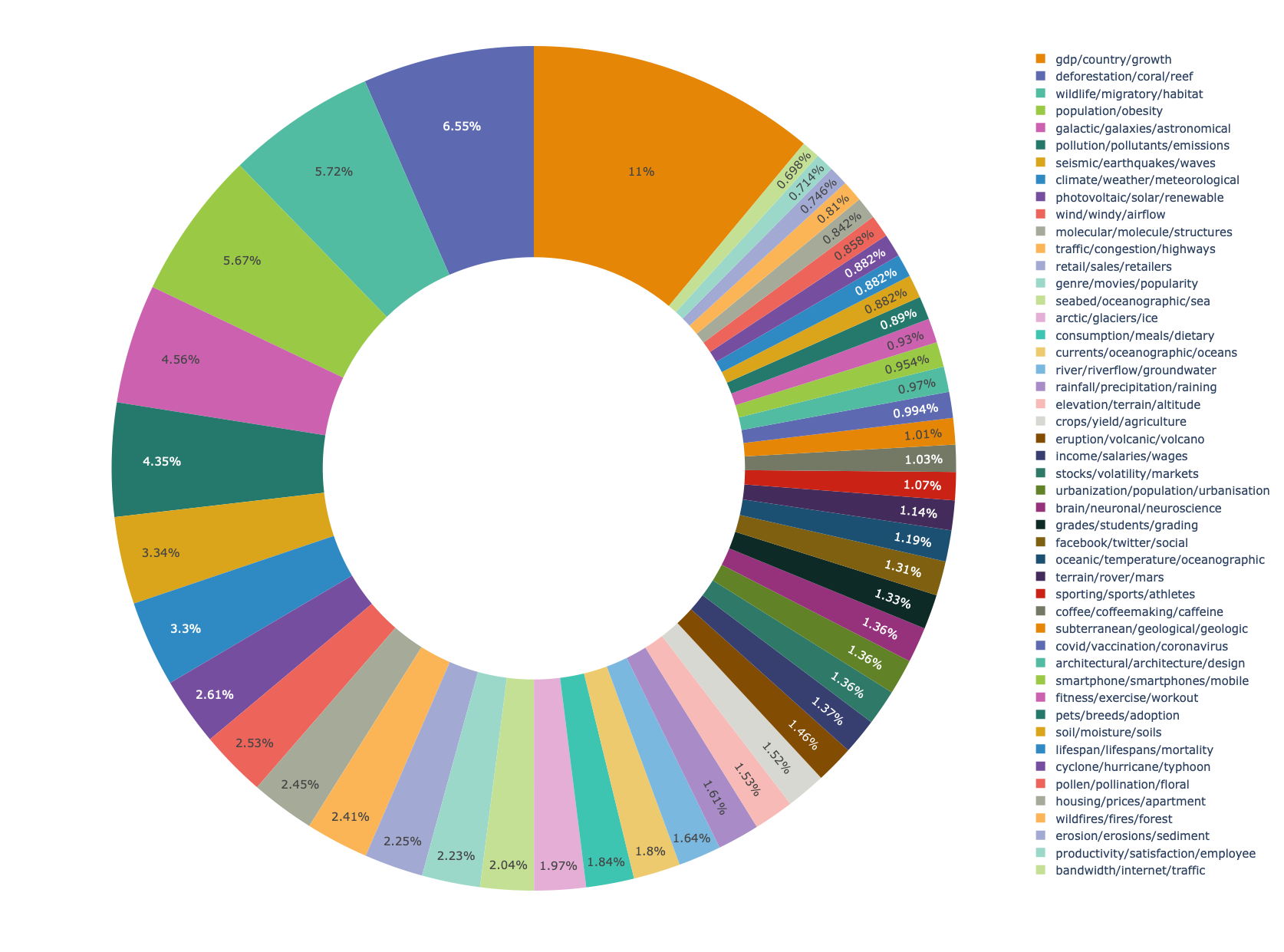}
    \vspace{1mm}
    \caption{Distribution of keywords within the topic pool, showcasing the diverse and balanced coverage of topics in the \ours{} dataset. }
    \label{fig:topic-dist}
\end{figure}

\Cref{fig:topic-dist} shows the distribution of keywords within the topic pool extracted using BERTopic\citep{grootendorst2022bertopic}. The generated topic pool encompasses a diverse range of fact-based and natural topics, ensuring comprehensive coverage across various subject areas.

\clearpage

\subsection{Plot Type Distribution}
\label{app:chart-type-dist}
\begin{figure}[ht]
\centering


\subfloat[\ours{} (ours)]{{\includegraphics[width=0.4\columnwidth ]{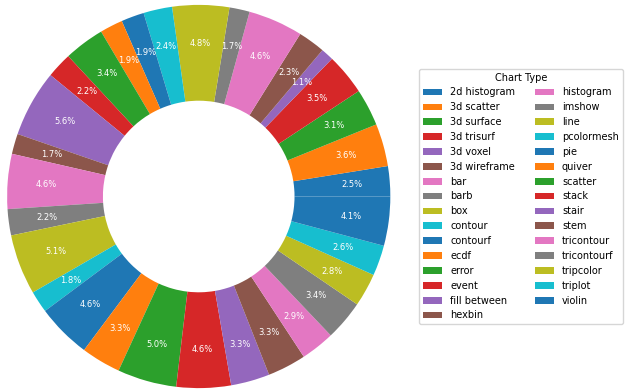} }}

\subfloat[PlotQA~\citep{methani2020plotqa}]{{\includegraphics[width=0.26\columnwidth ]
{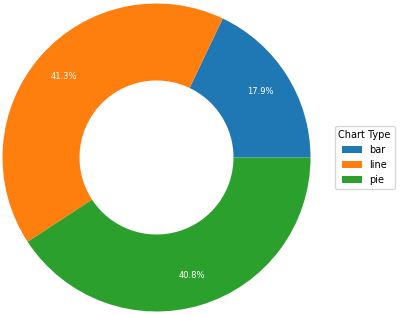} }} 
\hspace{0.02\columnwidth}
\subfloat[ChartQA~\cite{masry2022chartqa}]{{\includegraphics[width=0.26\columnwidth ]{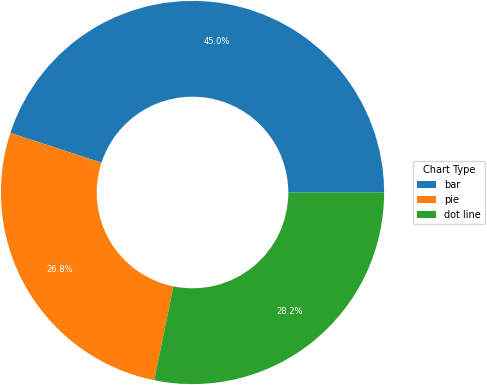} }} 
\hspace{0.02\columnwidth}
\subfloat[FigureQA~\cite{kahou2018figureqa}]{{\includegraphics[width=0.26\columnwidth ]{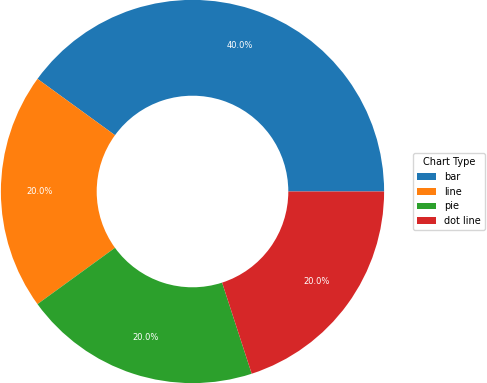} }} 
\hfill
\subfloat[Autochart~\cite{zhu2021autochart}]{{\includegraphics[width=0.31\columnwidth ]{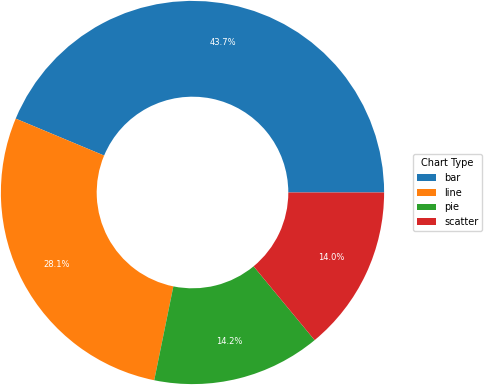} }} 
\hspace{0.02\columnwidth}
\subfloat[Chart-to-Text~\cite{kantharaj2022charttotext}]{{\includegraphics[width=0.31\columnwidth]{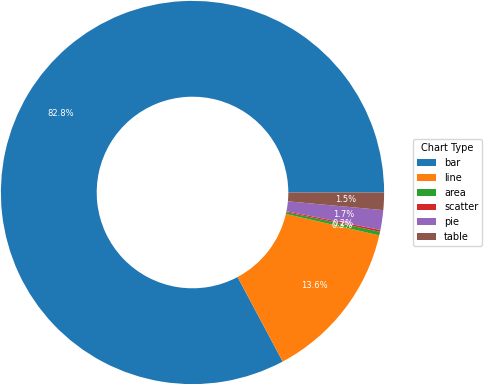} }} 
\hfill

\subfloat[ChartLlama~\cite{han2023chartllama}]{{\includegraphics[width=0.31\columnwidth ]{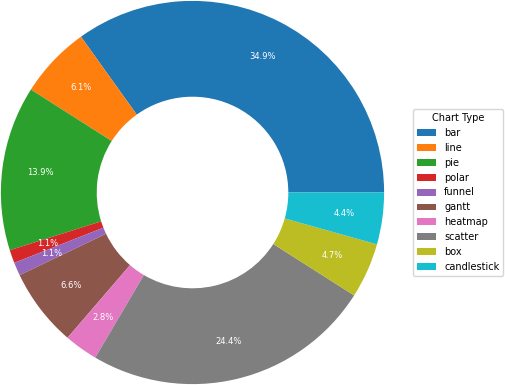} }} 
\hspace{0.02\columnwidth}
\subfloat[ChartX~\cite{xia2024chartx}]{{\includegraphics[width=0.31\columnwidth ]{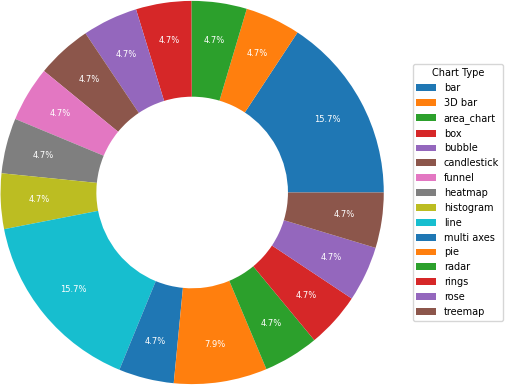} }} 
\hfill 

\caption{Comparison of the distribution of chart types with other datasets. Each pie chart shows the distribution of the plot types for each dataset, respectively. }
\label{fig:chart-type-dist}
\end{figure}

As shown in \Cref{fig:chart-type-dist}, our dataset, \ours{}, exhibits the most diverse and well-balanced distribution across various plot types when compared to other existing datasets. While existing datasets have primarily focused on common plot types such as bar charts and line charts, our dataset provides comprehensive coverage across a diverse range of plot types. This includes more complicated plot types like 3D surface plots and contour plots.

\clearpage



\subsection{Example of Text2Chart31}
\label{app:example-t4c31}

This section shows example from the Text2Chart31 dataset, providing a overall view of the data point including description, code, reasoning step, and data table, as shown in \Cref{tab:example}.

\begin{table}[ht]
\scriptsize
\centering
\begin{adjustbox}{max width=0.8\columnwidth}
    \begin{tabular}{@{}p{\columnwidth}@{}}
    \toprule
    \textbf{Description}\\
    \midrule
    This contour plot showcases the temperature variations along a coastal region using the dataset 'coastal\_temperature.csv'. The dataset includes information on latitude, longitude, and temperature (in degrees Celsius) for different coastal locations.

By establishing a grid of points (X, Y) representing latitude and longitude, we can model the temperature variations using the equation 'Z = a*X + b*Y + c', where a and b are regression coefficients determined through analysis.

Here are the first six rows of the dataset:\\
Latitude, Longitude, Temperature (°C)\\
40.7128, -74.0060, 15\\
48.8566, 2.3522, 10\\
37.7749, -122.4194, 18\\
51.5074, -0.1278, 12\\
35.6895, 139.6917, 20\\
42.3601, -71.0589, 14\\

Visualizing this data on a contour plot helps us analyze the spatial distribution of temperature along the coast. The contour lines indicate regions with similar temperature values, enabling us to identify temperature patterns and variations based on geographical location. This plot can be useful for understanding coastal climate, identifying areas with specific temperature ranges, and providing insights for planning activities related to beach tourism, coastal agriculture, and coastal development.
\\
    \midrule
    \textbf{Code}\\
    \midrule
...\\
\# Define the grid over which we will interpolate the data\\
x\_range = np.linspace(data['Longitude'].min(), data['Longitude'].max(), len(data['Longitude'].unique()))\\
y\_range = np.linspace(data['Latitude'].min(), data['Latitude'].max(), len(data['Latitude'].unique()))\\
x, y = np.meshgrid(x\_range, y\_range)\\
\\
\# Interpolate the measured temperature data onto the defined grid\\
z = griddata((data['Longitude'], data['Latitude']), data['Temperature (°C)'], (x, y), method='cubic')\\
\\
\# Configure the contour plot\\
plt.figure(figsize=(10, 6))\\
contours = plt.contourf(x, y, z, cmap='RdYlBu\_r')\\
plt.colorbar(contours)\\
... \\
    \midrule
    \textbf{Reasoning Step}\\
    \midrule
    1. Characteristics of the data and CSV file:\\
   - The data includes three columns: Latitude, Longitude, and Temperature (°C).\\
   - Latitude and Longitude are geographical coordinates specifying north-south and east-west positioning on the Earth's surface.\\
   - Temperature is a numerical measure indicating the amount of heat energy.\\
   - The data thus represents the temperature at different geographical locations around the world. ....\\
\\
2. Possible plot types:\\
   - **Map-based plots**: Utilizing the latitude and longitude values to plot geographical data on a global map.\\
   - **Scatter Plot**: To visualize how temperature varies with latitude or longitude.\\
   - **Contour or Heatmap**: Can be plotted over a map to show temperature distributions.\\
   - **3D Surface Plot**: Making use of X (Longitude), Y (Latitude), and Z (Temperature) to form a 3D visualization of temperature variations. \\
\\
3. Most suitable plot type:\\
   - **Map-based Heatmap or Contour Map**: This is especially suitable due to the geographical nature of the data (latitude and longitude) combined with a third variable (temperature). This type of visualization will allow viewers to easily understand temperature distributions across various geographical locations. ... \\
\\
4. Further considerations for the description:\\
   - Verify the accuracy and intent of the negative temperature values. If these are anomalies, it must be clearly stated to avoid misinterpretation.
   - Include proper color scales to represent temperature values, ensuring that the map is accessible and readable.\\
   - Consider adding interactive elements such as tooltips or zoom features if the visualization is digital, to enable detailed inspection of data at specific locations. ...\\

    \midrule
    \textbf{Data Table}\\
    \midrule
    Latitude,Longitude,Temperature (°C)\\
5.2927,-173.8335,31.8\\
-12.0503,72.5485,-39.1\\
86.2247,137.7667,27.6\\
-43.7762,-120.4625,44.5\\
16.645,-132.0993,-48.7\\
23.3972,-46.3823,-9.7\\
... \\
    \bottomrule

    \end{tabular}
    \end{adjustbox}   
\caption{An example data point in Text2Chart31, comprising a description, code, reasoning step, and CSV data table (top to bottom). The description elucidates the contour plot, coastal\_temperature.csv dataset, and insights from the visualization. The code utilizes Matplotlib for generating the contour plot. The reasoning step delineates the rationale behind crafting the data table and visualization, factoring in data characteristics, plot types, and additional consideration. Finally, the data table shows the dataset columns: Latitude, Longitude, and Temperature (°C).}
\label{tab:example}
\end{table}

\clearpage

\section{Experimental Details}
\label{app:detail}
\paragraph{Training setup and hyperparameters}
We report the hyperparameters for training supervised fine-tuning and joint reinforcement learning based fine-tuning in \Cref{tab:sft-detail} and \Cref{tab:rl-detail}.
For supervised fine-tuning, we fine-tune base model with LoRA adapter with the configuration in \Cref{tab:sft-detail}.
For reinforcement learning-based fine-tuning, we start with the supervised fine-tuned model and merge the LoRA parameters into the original model parameters. 
Then, we apply an additional LoRA adapter according to the configuration in Table \ref{tab:rl-detail}. Finally, we fine-tune both Task 1 and Task 3 models jointly using the PPO algorithm.

\begin{table}[ht]
    \centering
    \begin{adjustbox}{max width=1.0\columnwidth}
    \begin{tabular}{lccccc}
    \toprule
     & \multicolumn{2}{c}{Task 1} & \multicolumn{2}{c}{Task 2} & Task 3 \\
     \cmidrule(r){2-3} \cmidrule(r){4-5} \cmidrule(r){6-6}
    Model & L3I-8B & CLI-13B & L2C-7B & L3I-8B & L3I-8B\\
    \midrule
    Training epochs & 5 & 5 & 5 & 5 & 5 \\
    Training set size & 9705 & 9705 & 7142 & 7142 & 9705 \\
    Batch size & 16 & 16 & 16 & 16 & 16  \\
    Optimizer & AdamW & AdamW & AdamW & AdamW & AdamW \\
    Learning rate & 5e-4 & 5e-4 & 5e-5 & 5e-5 & 5e-4 \\
    Learning rate scheduling & Constant & Constant & Constant & Constant & Constant \\
    Mixed precision & BF16 & BF16 & BF16 & BF16 & BF16 \\
    LoRA rank & 16 & 16 & 32 & 32 & 32\\
    LoRA alpha & 16 & 16 & 32 & 32 & 32\\
    LoRA dropout & 0.1 & 0.1 & 0.1 & 0.1 & 0.1\\
    \bottomrule

    \end{tabular}
    \end{adjustbox}
\caption{Training hyperparameters for supervised fine-tuning. L3I-8B, CLI-13B, and L2C-7B denote Llama 3 Instruct-8B, Code Llama Instruct-13B, and Llama 2 Chat-7B, respectively.}
    \label{tab:sft-detail}
\end{table}

\begin{table}[ht]
    \centering

    \begin{adjustbox}{max width=1.0\columnwidth}
    \begin{tabular}{lccc}
    \toprule
     & \multicolumn{2}{c}{Task 1}& Task 3 \\
     \cmidrule(r){2-3} \cmidrule(r){4-4} 
    Model & L3I-8B & CLI-13B & L3I-8B \\
    \midrule
    Batch size & 8 & 8 & 8   \\
    Training steps & 63 & 94 & 63 \\
    Training data size & 504 & 752 & 504 \\
    Optimizer & Adam  & Adam  &  Adam \\
    Learning rate & 1.41e-5 &  7.05e-6 & 1.41e-5 \\
    Learning rate scheduling & Constant & Constant & Constant \\
    Mixed precision & BF16 & BF16 & BF16 \\
    LoRA rank & 16 & 16 & 32 \\
    LoRA alpha & 16 & 16 & 32 \\
    LoRA dropout & 0.1 & 0.1 & 0.1 \\
    \bottomrule

    \end{tabular}
    \end{adjustbox}
\caption{Training hyperparameters for RL fine-tuning. L3I-8B and CLI-13B denote Llama 3 Instruct-8B and Code Llama Instruct 13B, respectively.}
\label{tab:rl-detail}
\end{table}

\clearpage

\section{Cycle Consistency Details}
\label{detail:cycle}
\begin{figure}[ht]
    \centering
    \includegraphics[width=0.6\columnwidth]{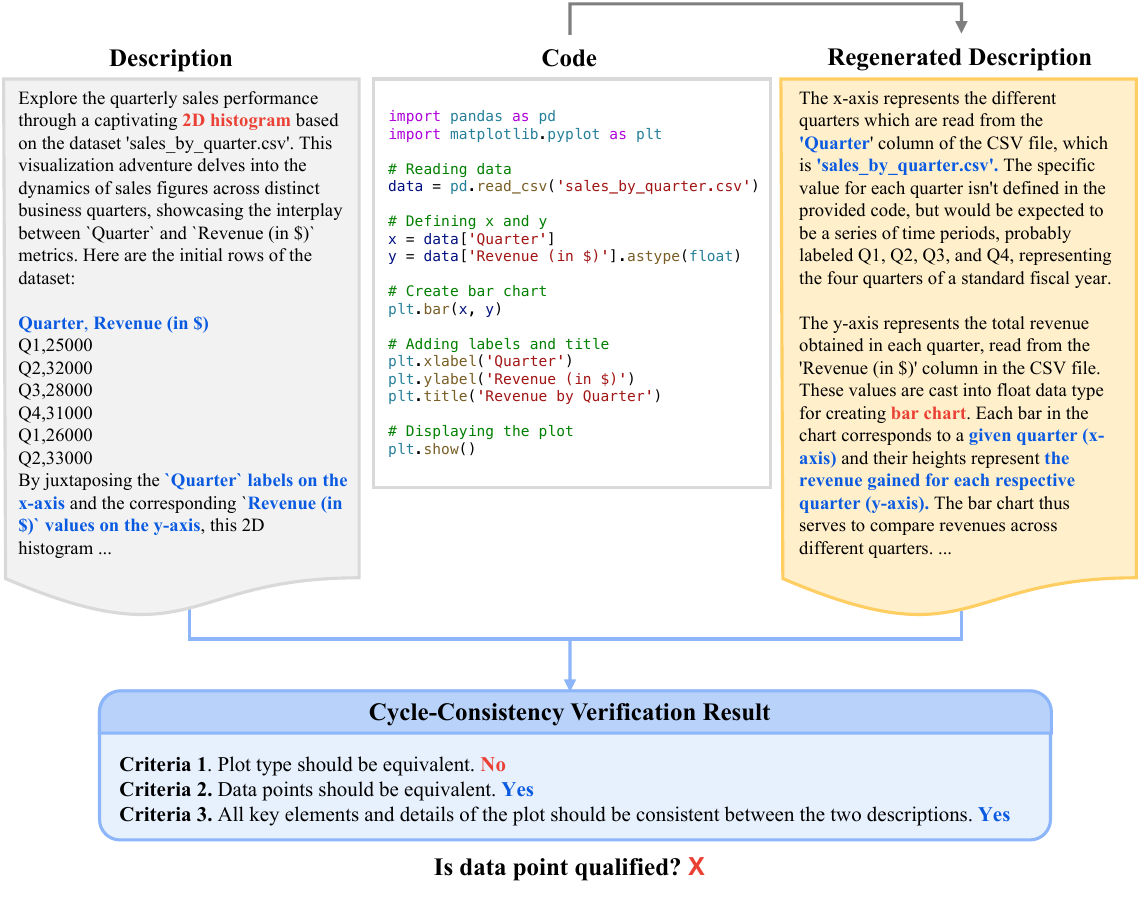}
    \caption{Example of cycle consistency verification for a description and generated code, showcasing inconsistency in the plot type (2D histogram vs. bar chart) despite consistent data source and sufficient detail in both descriptions.}
    \label{fig:cycle-consistency-disqualified}
\end{figure}

\begin{figure}[ht]
    \centering
    \includegraphics[width=0.6\columnwidth]{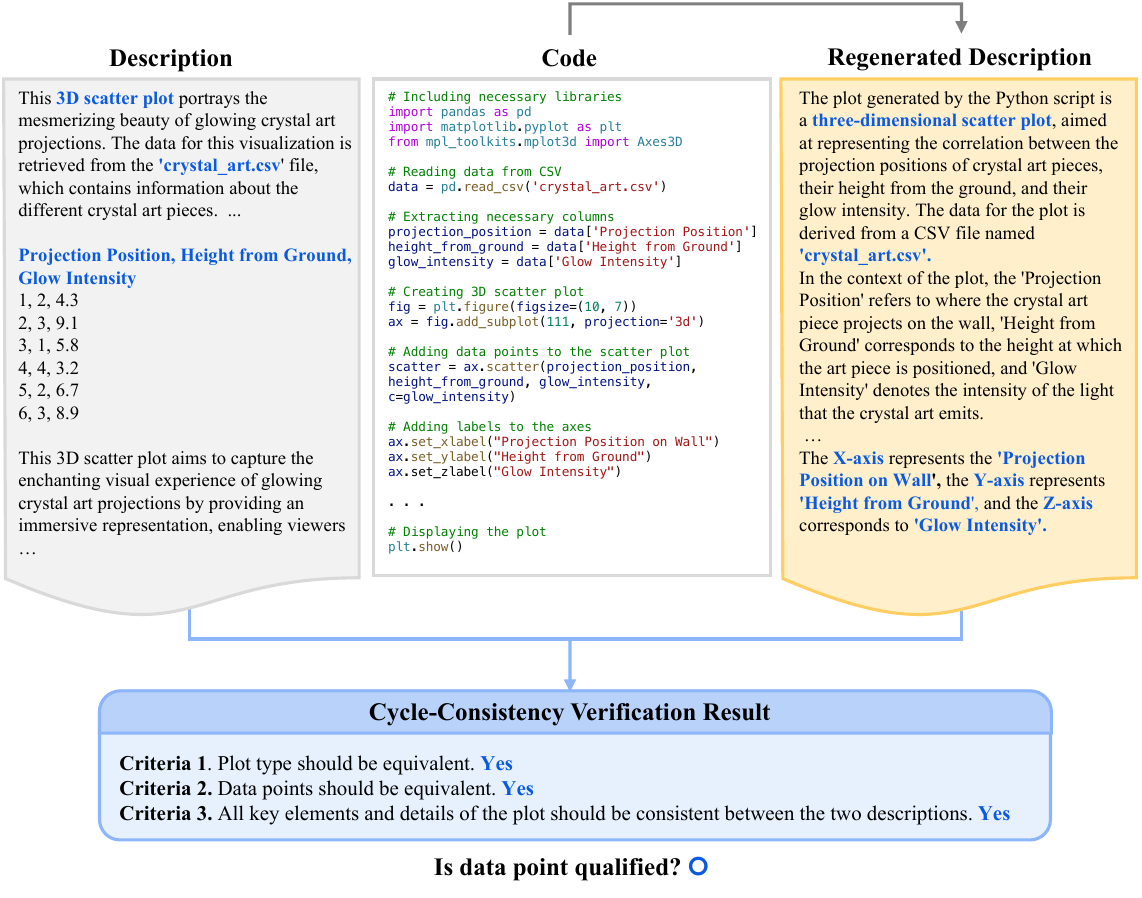}
    \caption{Example of cycle consistency verification for a description and generated code, showcasing consistency in the plot type (3D scatter plot), data source (crystal\_art.csv), and sufficient detail in both descriptions to accurately redraw the plot.}
    \label{fig:cycle-consistency-qualified}
\end{figure}

This method leverages the capabilities of language models to verify the consistency between the original plot description and the generated code, without the need for manual human evaluation. \Cref{fig:cycle-consistency-disqualified} and \Cref{fig:cycle-consistency-qualified} illustrate examples of data points that fail and pass the cycle consistency verification, respectively. By employing this method, we ensure that the generated code and plot are well aligned with the intended visualization described in the original description, maintaining the quality of the \ours{} dataset.

\clearpage
\section{Prompt Template}
\label{app:prompt}
\subsection{Prompt Template used for Data Generation}
\begin{figure}[ht]
    \centering
    \includegraphics[width=0.75\columnwidth]{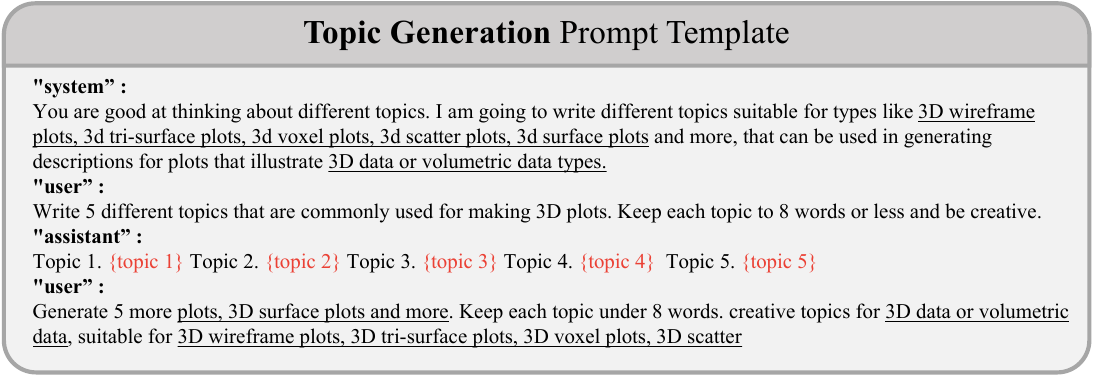}
    \caption{Prompt template used for topic generation}
    \label{fig:tem-topic}
\end{figure}

\begin{figure}[ht]
    \centering
    \includegraphics[width=0.75\columnwidth]{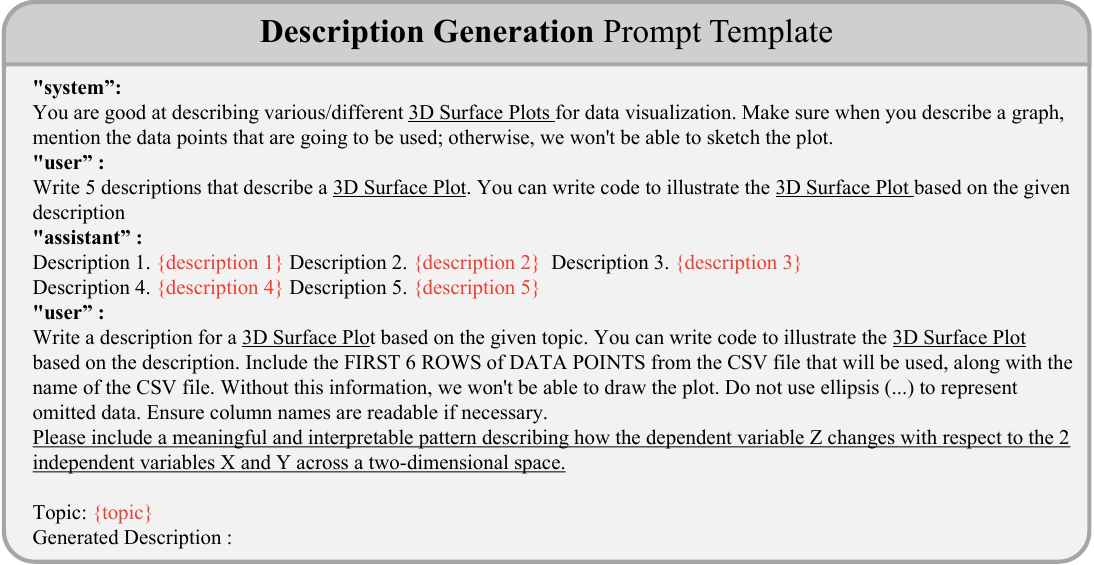}
    \caption{Prompt template used for description generation}
    \label{fig:tem-des}
\end{figure}

\begin{figure}[ht]
    \centering
    \includegraphics[width=0.75\columnwidth]{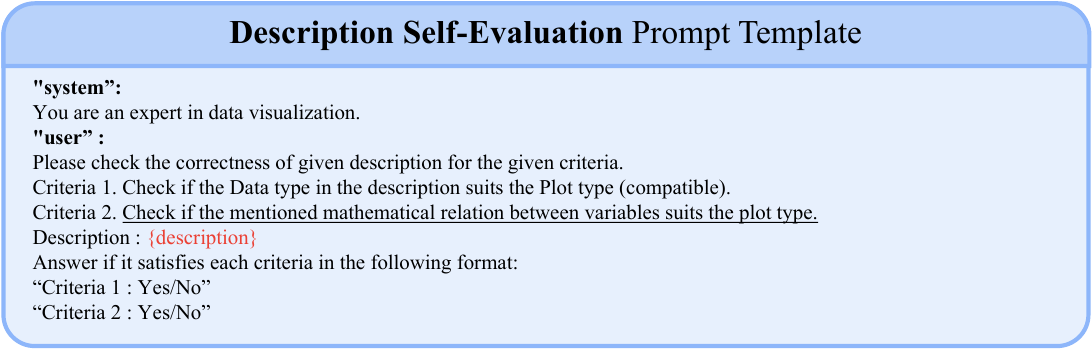}
    \caption{Prompt template used for description self-evalution}
    \label{fig:tem-self-eval}
\end{figure}

\begin{figure}[ht]
    \centering
    \includegraphics[width=0.75\columnwidth]{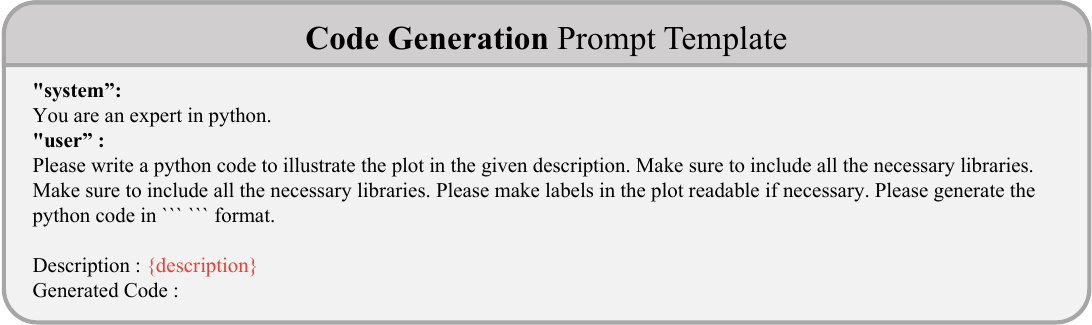}
    \caption{Prompt template used for code generation}
    \label{fig:tem-code}
\end{figure}

\begin{figure}[ht]
    \centering
    \includegraphics[width=0.75\columnwidth]{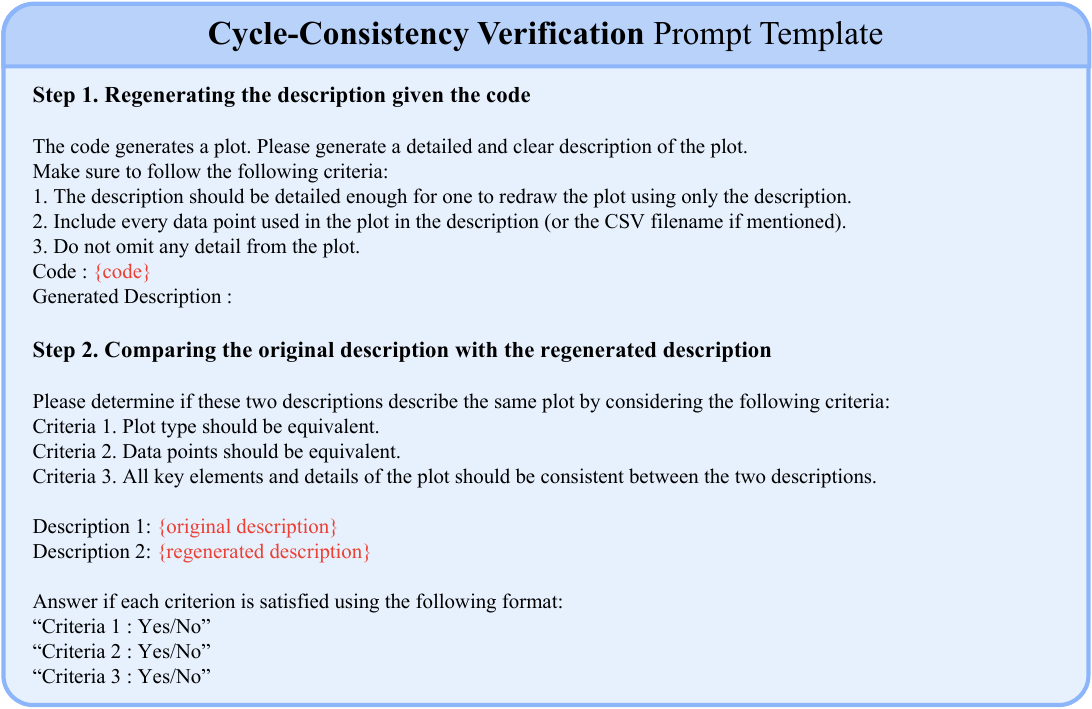}
    \caption{Prompt template used for cycle consistency verification }
    \label{fig:tem-cycle-consistency}
\end{figure}

\begin{figure}[ht]
    \centering
    \includegraphics[width=0.74\columnwidth]{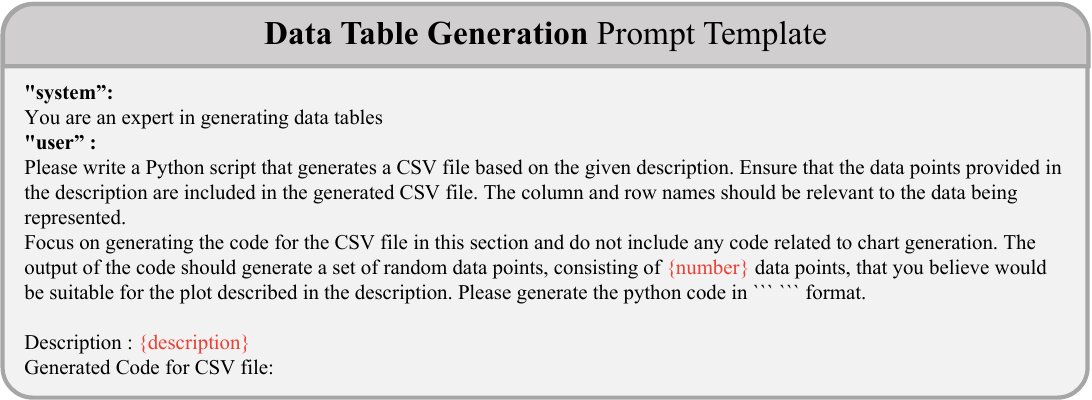}
    \caption{Prompt template used for data table generation}
    \label{fig:tem-table}
\end{figure}

\begin{figure}[ht]
    \centering
    \includegraphics[width=0.75\columnwidth]{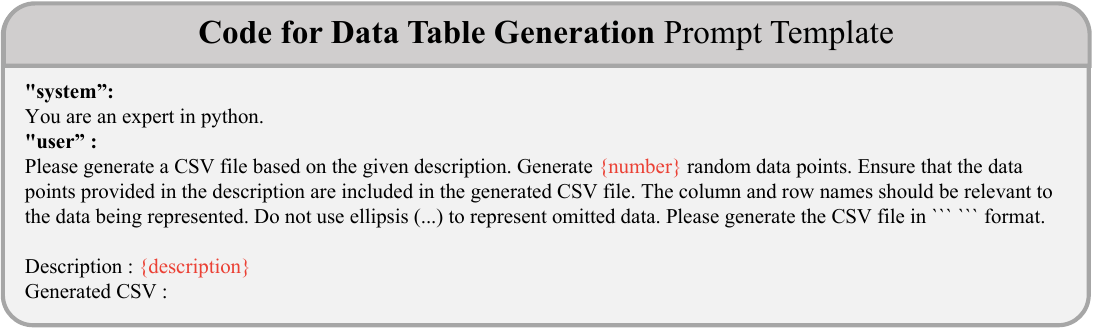}
    \caption{Prompt template used for code used for data table generation}
    \label{fig:tem-code-table}
\end{figure}

\begin{figure}[ht]
    \centering
    \includegraphics[width=0.75\columnwidth]{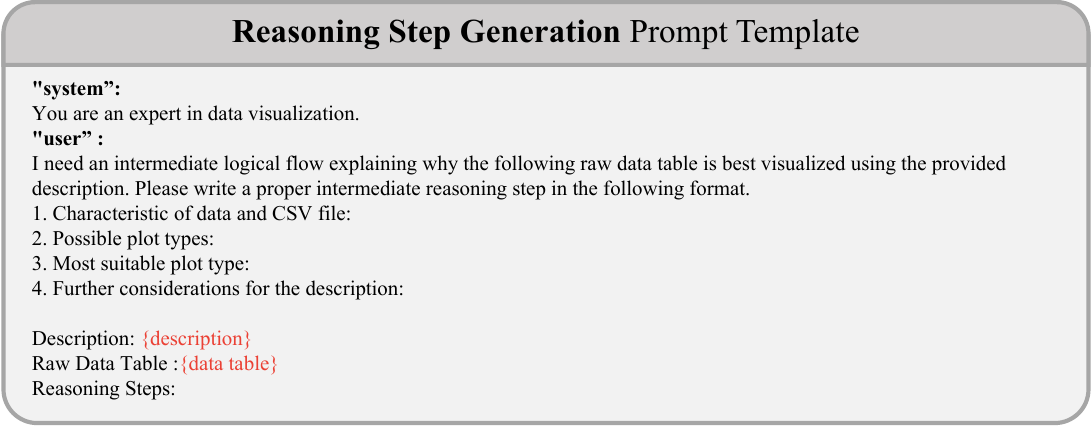}
    \caption{Prompt template used for reasoning step generation}
    \label{fig:tem-reasoning-step}
\end{figure}

This section presents the prompt templates used for various stages of the data generation process in the \ours{} dataset. 
\Cref{fig:tem-topic} illustrates the prompt template used for topic generation, while \Cref{fig:tem-des} shows the template for description generation. \Cref{fig:tem-self-eval} presents the template for description self-evaluation, and \Cref{fig:tem-code} illustrates the template for code generation. \Cref{fig:tem-cycle-consistency} shows the template used for cycle consistency verification, and \Cref{fig:tem-table} presents the template for data table generation. \Cref{fig:tem-code-table} illustrates the template for generating code that creates the data table, and \Cref{fig:tem-reasoning-step} shows the template for reasoning step generation.

\vspace{3mm}

\clearpage
\subsection{Prompt Template for Tasks}
\begin{figure}[ht]
    \centering
    \includegraphics[width=0.75\columnwidth]{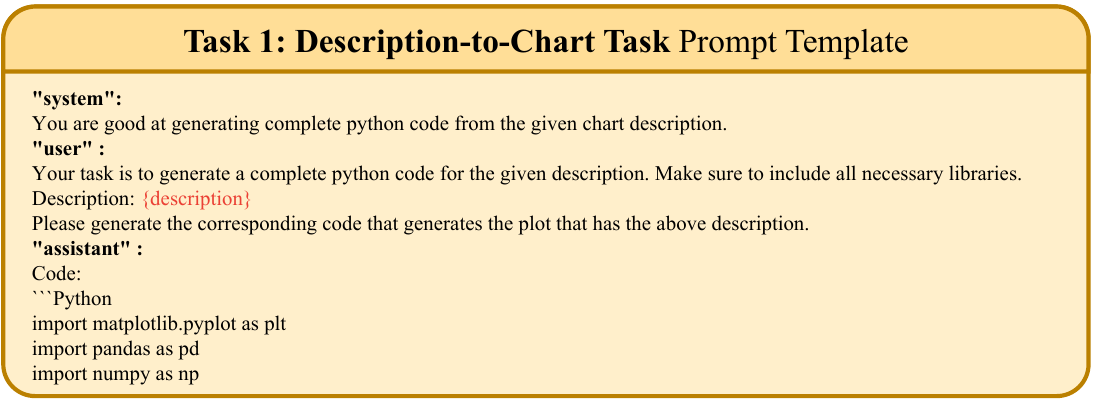}
    \caption{Prompt template used for description to chart task}
    \label{fig:tem-task1}
\end{figure}

\begin{figure}[ht]
    \centering
    \includegraphics[width=0.75\columnwidth]{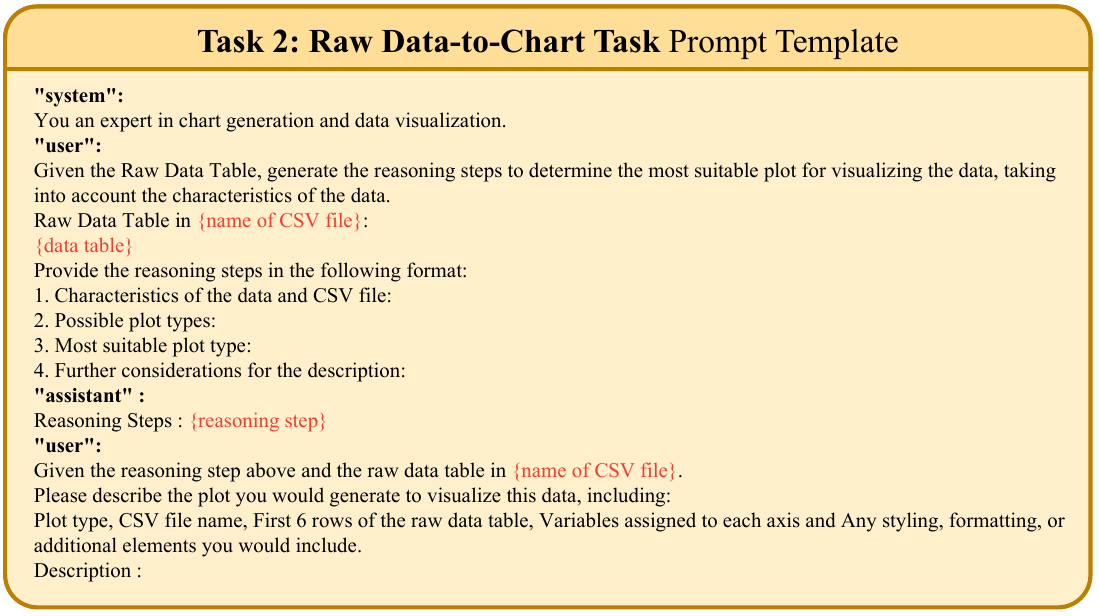}
    \caption{Prompt template used for raw data-to-chart task}
    \label{fig:tem-task2}
\end{figure}

\begin{figure}[ht]
    \centering
    \includegraphics[width=0.75\columnwidth]{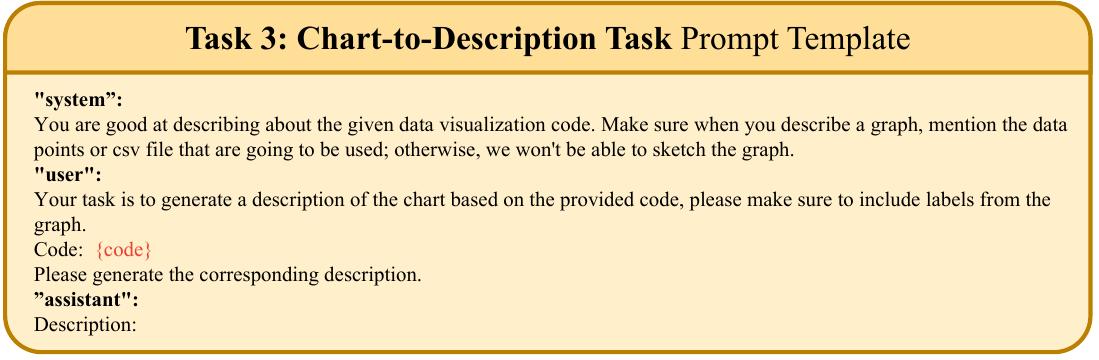}
    \caption{Prompt template used for chart-to-description task }
    \label{fig:tem-task3}
\end{figure}

This section presents the prompt templates used for three tasks using the \ours{} dataset, including description-to-chart, raw data-to-chart, and chart-to-description tasks.
\Cref{fig:tem-task1} illustrates the prompt template used for the description-to-chart task, \Cref{fig:tem-task2} shows the template for the raw data-to-chart task, and \Cref{fig:tem-task3} presents the template used for the chart-to-description task.

\clearpage

\section{Details on Human Evaluation}
\label{app:human-eval}
\begin{figure}[ht]
    \centering
    \includegraphics[width=1.0\columnwidth]{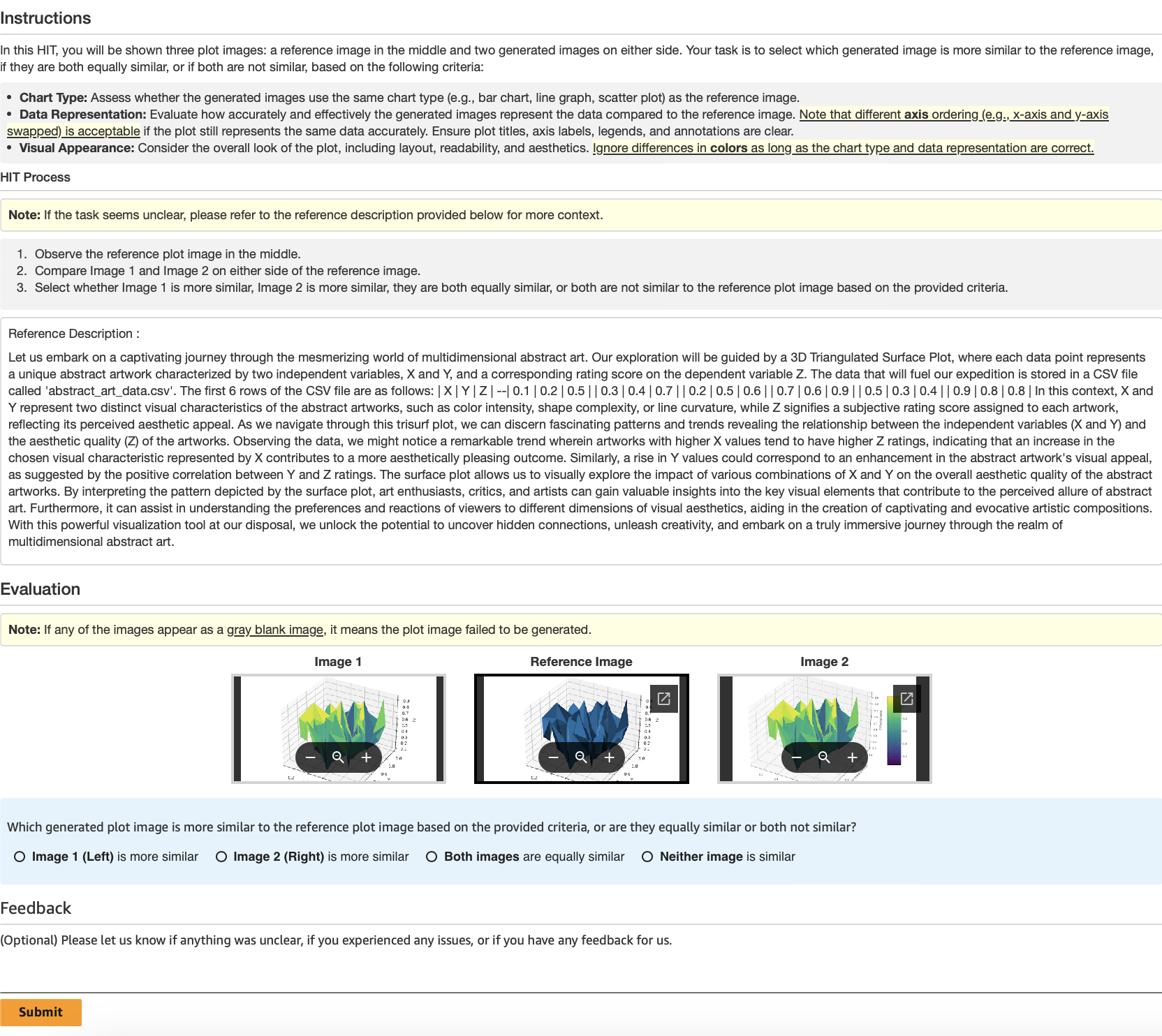}
    \caption{User interface for human evaluation comparing generated plot images.}
    \label{fig:human-eval-UI}
\end{figure}

\Cref{fig:human-eval-UI} illustrates the user interface designed for the human evaluation task. The interface presents the crowd workers with a reference plot image and two generated plot images (Image 1 and Image 2) from different models, where the order of the generated images is randomly determined. The workers are asked to select one of the following options: Image 1 (Left) is more similar to the reference image, Image 2 (Right) is more similar to the reference image, both images are equally similar to the reference image, or neither image is similar to the reference image. The workers make their selection based on the similarity of the generated images to the reference image in terms of chart type, data representation, and visual appearance. We use Amazon Mechanical Turk and gather annotators from English speaking countries. We pay maximum \$0.4 per HIT. We explain annotators that the provided answers are going to be used as a research purpose in our qualification HIT.

\clearpage

\end{document}